\documentclass[table,xcdraw]{ai2style/ai2}

\usepackage{tabularx}
\usepackage{lineno}
\usepackage{enumitem}
\usepackage{listings}
\usepackage{svg}
\usepackage{amssymb}
\usepackage{multirow}
\usepackage{bigdelim}
\usepackage{todonotes}
\usepackage{longtable}
\usepackage{tabularray}
\usepackage{wrapfig}
\usepackage{xspace}
\usepackage[absolute]{textpos}
\usepackage{fdsymbol}
\usepackage[utf8]{inputenc}
\usepackage[T1]{fontenc}
\usepackage{amsfonts}
\usepackage{nicefrac}
\usepackage{amsmath}
\usepackage{csquotes}
\usepackage{tablefootnote}
\usepackage{siunitx}
\usepackage{arydshln}
\usepackage{soul}

\newcommand{\ranjay}[1]{{\color{orange}$[$#1$]^R_K$}}
\newcommand{\TODO}[1]{\textbf{\color{red}[TODO: #1]}}
\newcommand{\jr}[1]{\textbf{\color{cyan}[JR: #1]}}
\newcommand{\wk}[1]{\textbf{\color{green}[WK: #1]}}

\definecolor{baselinecolor}{gray}{.9}

\renewcommand{\TODO}[1]{}

\renewcommand{\jr}[1]{}
\renewcommand{\ranjay}[1]{}
\renewcommand{\wk}[1]{}

\usepackage{multirow}
\usepackage{graphicx}
\usepackage{listings}
\lstdefinelanguage{Prompt}{
  morestring=[b]",
}
\definecolor{codebg}{RGB}{245,245,245}
\usepackage{adjustbox}
\usepackage{pifont}
\usepackage{caption}
\usepackage{makecell}
\usepackage{subcaption}
\usepackage{bold-extra}
\usepackage{pgf-pie}

\definecolor{linkcolor}{RGB}{0, 0, 128}
\hypersetup{
     colorlinks   = true,
     citecolor    = linkcolor,
     linkcolor    = linkcolor,
     urlcolor     = linkcolor,
}

\setlist[itemize]{leftmargin=*,itemsep=0em,parsep=0.3em,topsep=0.3em}

\DeclareUnicodeCharacter{2212}{\ensuremath{-}}

\definecolor{maroon}{HTML}{F26035}
\definecolor{yellow}{HTML}{FDBC42}
\definecolor{lavender}{HTML}{734f96}
\definecolor{darkergrey}{HTML}{444444}
\definecolor{midgrey}{HTML}{e6eded}
\definecolor{ai2pink}{HTML}{f0529c}
\definecolor{ai2midpink}{HTML}{fad3e5}
\definecolor{ai2lightpink}{HTML}{fbecf3}
\definecolor{ai2midwhite}{HTML}{f2e5d9}
\definecolor{ai2offwhite}{HTML}{fbf4ee}
\definecolor{ai2green}{HTML}{0fcb8c}
\definecolor{ai2lightgreen}{HTML}{e7f9f3}
\definecolor{ai2darkgreen}{HTML}{105257}
\definecolor{ai2purple}{HTML}{B932EB}
\definecolor{ai2lightpurple}{HTML}{f7e8fc}
\definecolor{neutralEight}{HTML}{343434}
\definecolor{neutralFive}{HTML}{838383}
\definecolor{neutralThree}{HTML}{bebebe}
\definecolor{neutralOne}{HTML}{dedede}
\definecolor{lightgrey}{HTML}{fafcfc}
\definecolor{plum}{rgb}{0.56,0.27,0.52}
\definecolor{ai2blue}{HTML}{b3dadf}

\usepackage{tikz}

\definecolor{maroon}{HTML}{F26035}
\definecolor{yellow}{HTML}{FDBC42}
\definecolor{darkred}{RGB}{156, 39, 33}
\definecolor{darkblue}{RGB}{31, 90, 153}
\definecolor{forestgreen}{rgb}{0.13, 0.55, 0.13}
\definecolor{brickred}{rgb}{0.8, 0.25, 0.33}
\definecolor{olmoDarkBlue}{HTML}{012e59}
\definecolor{olmoBlue}{HTML}{265ed4}
\definecolor{olmoLightBlue}{HTML}{012e59}
\definecolor{olmoTeal}{HTML}{00d5ff}
\definecolor{olmoYellow}{HTML}{ffbb00}
\definecolor{olmoOrange}{HTML}{ff9100}

\newcommand{\app}{\raise.17ex\hbox{$\scriptstyle\sim$}}
\renewcommand{\paragraph}[1]{\vspace{0.5mm}\noindent\textbf{#1}}
\newcommand{\tablestyle}[2]{\setlength{\tabcolsep}{#1}\renewcommand{\arraystretch}{#2}\centering\footnotesize}

\definecolor{molmocolor}{RGB}{240, 82, 156}
\definecolor{tablegray}{RGB}{223, 242, 252}
\definecolor{tablegreen}{RGB}{15, 203, 150}
\definecolor{tableyellow}{RGB}{250, 242, 233}
\definecolor{tableblue}{RGB}{240, 82, 156}
\definecolor{darkpink}{RGB}{139, 14, 98}

\usepackage{setspace}

\usepackage{nicematrix}
\newcolumntype{L}[1]{>{\raggedright\let\newline\\\arraybackslash\hspace{0pt}}m{#1}}
\newcolumntype{C}[1]{>{\centering\let\newline\\\arraybackslash\hspace{0pt}}m{#1}}
\newcolumntype{R}[1]{>{\raggedleft\let\newline\\\arraybackslash\hspace{0pt}}m{#1}}
\newcolumntype{P}[1]{>{\centering\let\newline\\\arraybackslash\columncolor{ai2lightpink}}m{#1}}
\newcommand{\huggingface}{\raisebox{-1.5pt}{\includegraphics[height=1.05em]{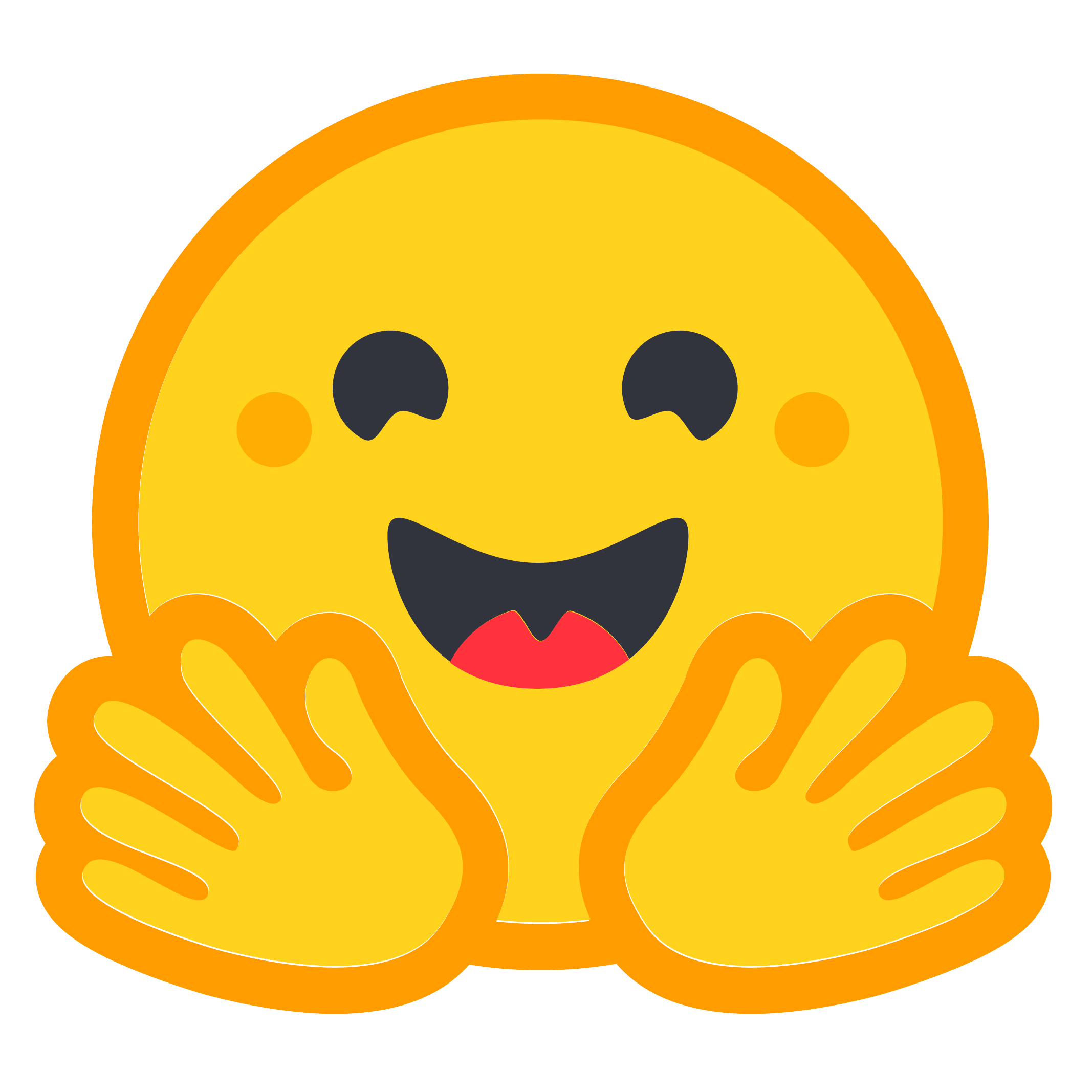}}\xspace}
\newcommand{\hfdataset}{\raisebox{-1.5pt}{\includegraphics[height=1.05em]{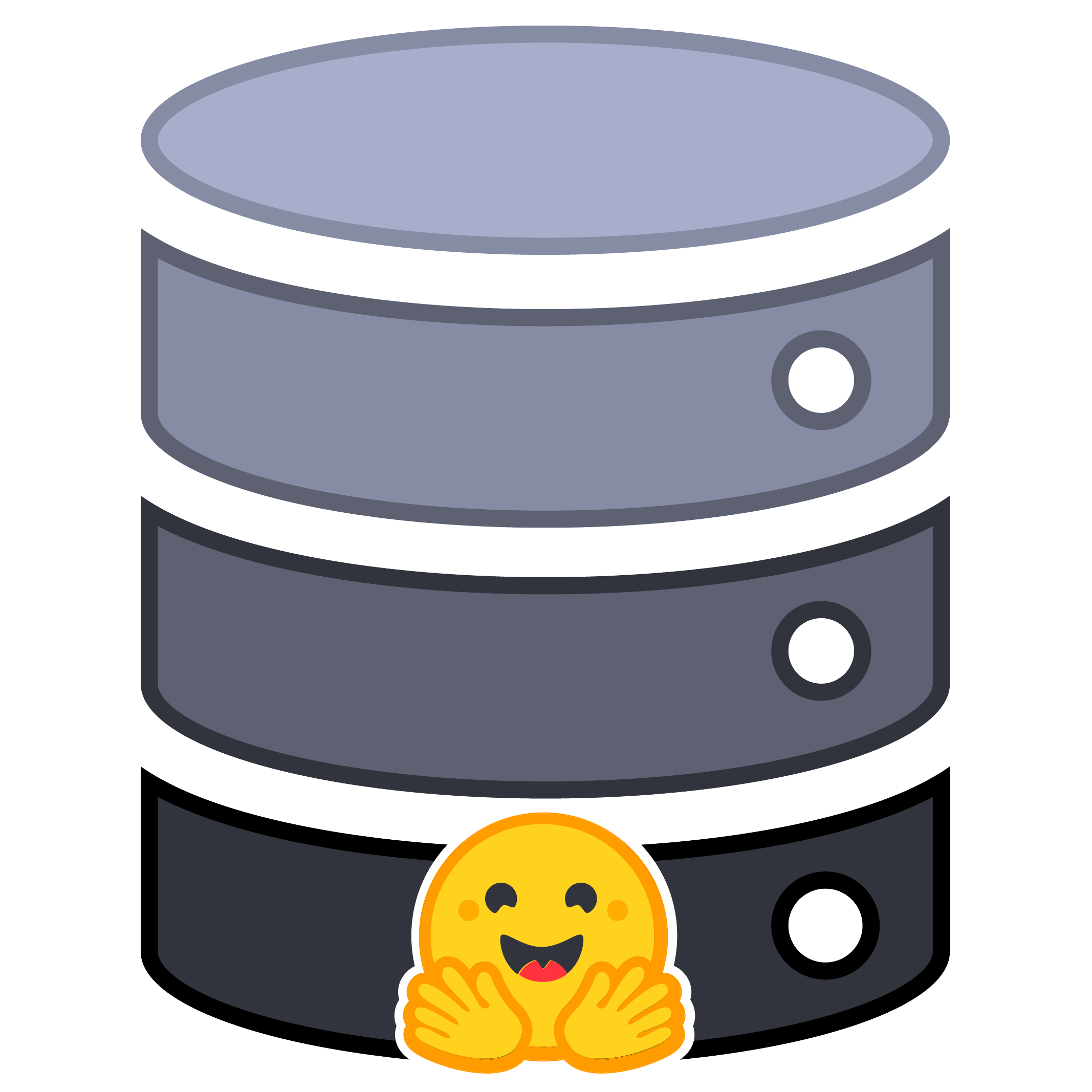}}\xspace}

\newcommand{\github}{\raisebox{-1.5pt}{\includegraphics[height=1.05em]{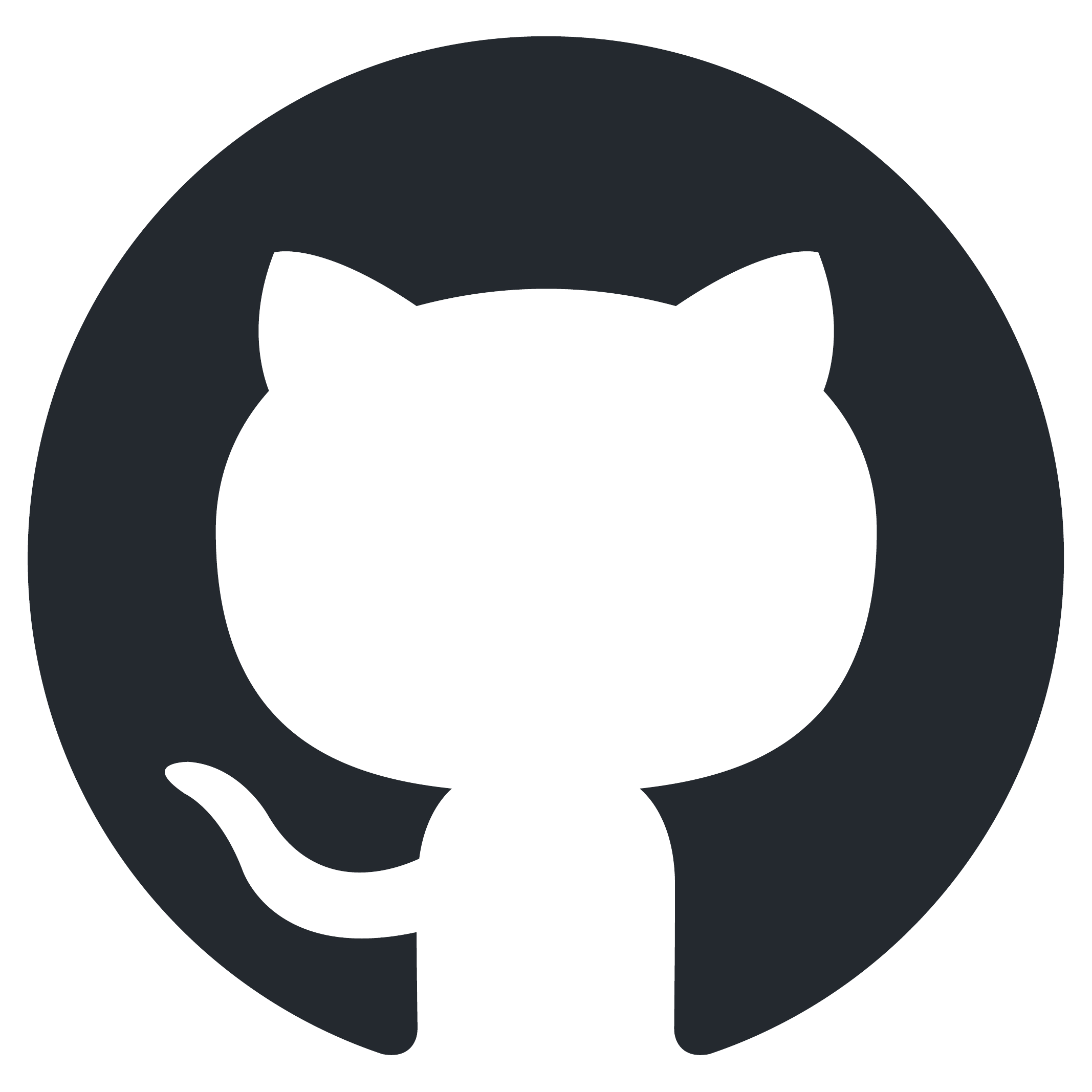}}\xspace}

\makeatletter
\DeclareRobustCommand\onedot{\futurelet\@let@token\@onedot}
\def\@onedot{\ifx\@let@token.\else.\null\fi\xspace}
\def\eg{\emph{e.g}\onedot}

\def\ie{\emph{i.e}\onedot}

\def\vs{\emph{vs}\onedot}

\def\etal{\emph{et al}\onedot}
\makeatother

\title{
\mbox{\includegraphics[width=0.65cm]{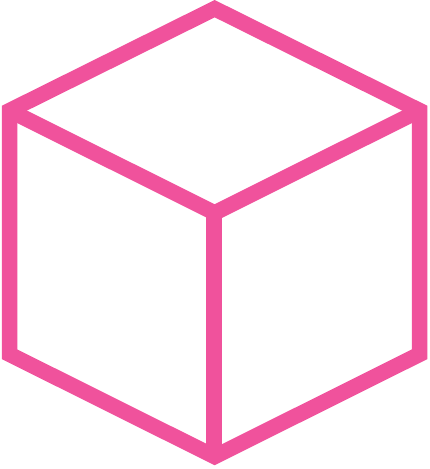}} 
WildDet3D \\ {\fontsize{18pt}{12pt}\selectfont Scaling Promptable 3D Detection in the Wild}
}

\newcommand{\core}{\textsuperscript{\textcolor{ai2pink}{\ding{170}}}}

\authorOne[1,2]{Weikai Huang\core}
\authorOne[1,2]{Jieyu Zhang\core}

\authorTwo[2]{Sijun Li}
\authorTwo[2]{Taoyang Jia}
\authorTwo[1,2]{Jiafei Duan}
\authorTwo[1]{Yunqian Cheng}
\authorTwo[1,2,5]{Jaemin Cho}
\authorTwo[1]{Matthew Wallingford}
\authorTwo[1,2]{Rustin Soraki}
\authorTwo[1]{Chris Dongjoo Kim}
\authorTwo[1,2]{Shuo Liu}
\authorTwo[1,2]{Donovan Clay}
\authorTwo[1]{Taira Anderson}
\authorTwo[1]{Winson Han}
\authorThree[1,2]{Ali Farhadi}
\authorThree[3]{Bharath Hariharan}
\authorThree[1,2,4]{Zhongzheng Ren\core}
\authorThree[1,2]{Ranjay Krishna\core}

\affiliation[1]{Allen Institute for AI}
\affiliation[2]{University of Washington}
\affiliation[3]{Cornell University}
\affiliation[4]{UNC-Chapel Hill}
\affiliation[5]{Johns Hopkins University}

\contribution[]{$\textcolor{ai2pink}{\varheartsuit}$ marks core contributors.}

\newcommand{\model}{WildDet3D\xspace}
\newcommand{\data}{WildDet3D-Data\xspace}
\newcommand{\bench}{WildDet3D-Bench\xspace}

\definecolor{tablegray}{RGB}{223, 242, 252}
\definecolor{tablegreen}{RGB}{15, 203, 150}
\definecolor{tableyellow}{RGB}{250, 242, 233}
\definecolor{tableblue}{RGB}{240, 82, 156}
\definecolor{darkpink}{RGB}{139, 14, 98}
\definecolor{baselinecolor}{gray}{.9}

\abstract{
Understanding objects in 3D from a single image is a cornerstone of spatial intelligence. A key step toward this goal is monocular 3D object detection—recovering the extent, location, and orientation of objects from an input RGB image. To be practical in the open world, such a detector must generalize beyond closed-set categories, support diverse prompt modalities, and leverage geometric cues when available. Progress is hampered by two bottlenecks: existing methods are designed for a single prompt type and lack a mechanism to incorporate additional geometric cues, and current 3D datasets cover only narrow categories in controlled environments, limiting open-world transfer.
In this work we address both gaps. First, we introduce \textbf{\model}, a unified geometry-aware architecture that natively accepts text, point, and box prompts and can incorporate auxiliary depth signals at inference time. Second, we present \textbf{\data}, the largest open 3D detection dataset to date, constructed by generating candidate 3D boxes from existing 2D annotations and retaining only human-verified ones, yielding over 1M images across 13.5K categories in diverse real-world scenes.
\model establishes a new state-of-the-art across multiple benchmarks and settings. 
In the open-world setting, it achieves 22.6/24.8 AP$_\text{3D}$ on our newly introduced \textbf{\bench} with text and box prompts.
On Omni3D, it reaches 34.2/36.4 AP$_\text{3D}$ with text and box prompts, respectively.
In zero-shot evaluation, it achieves 40.3/48.9 ODS on Argoverse~2 and ScanNet.
 Notably, incorporating depth cues at inference time yields substantial additional gains (+20.7 AP on average across settings).
}

\metadata[\quad\huggingface Models:]{
\href{https://huggingface.co/allenai/WildDet3D}{\texttt{\model}}}

\metadata[\vspace{.5em}\quad\hfdataset Data:]{
    \href{https://huggingface.co/datasets/allenai/WildDet3D-Data}{\texttt{\data}} \quad \href{https://huggingface.co/datasets/allenai/WildDet3D-Bench}{\texttt{\bench}} \quad \href{https://huggingface.co/datasets/allenai/WildDet3D-Stereo4D-Bench}{\texttt{WildDet3D-Stereo4D-Bench}}
}

\metadata[\vspace{.5em}\quad\github Code:]{
    \href{https://github.com/allenai/WildDet3D}{\texttt{https://github.com/allenai/WildDet3D}} \quad
}

\metadata[\vspace{.5em}\quad\raisebox{-1.5pt}{\includegraphics[height=1.05em]{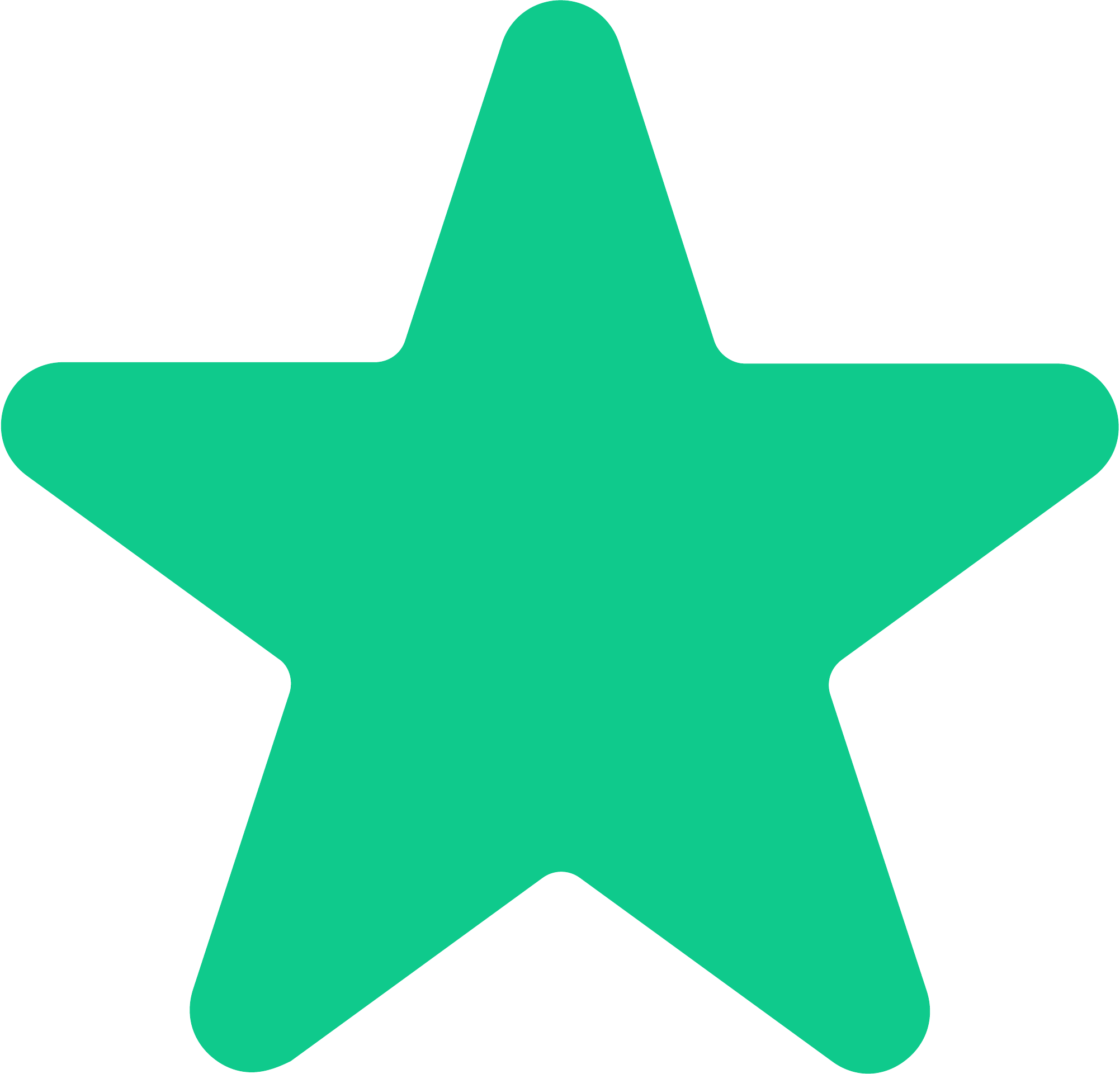}}\xspace Demo:]{\href{https://huggingface.co/spaces/allenai/WildDet3D}{\texttt{https://huggingface.co/spaces/allenai/WildDet3D}}}

\metadata[\vspace{.5em}\quad\raisebox{-1.5pt}{\includegraphics[height=1.05em]{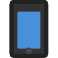}}\xspace iPhone App:]{\href{https://apps.apple.com/us/app/wilddet3d/id6760861157}{\texttt{\model~(App Store)}}}

\metadata[\vspace{.5em}\quad\raisebox{-1.5pt}{\includegraphics[height=1.05em]{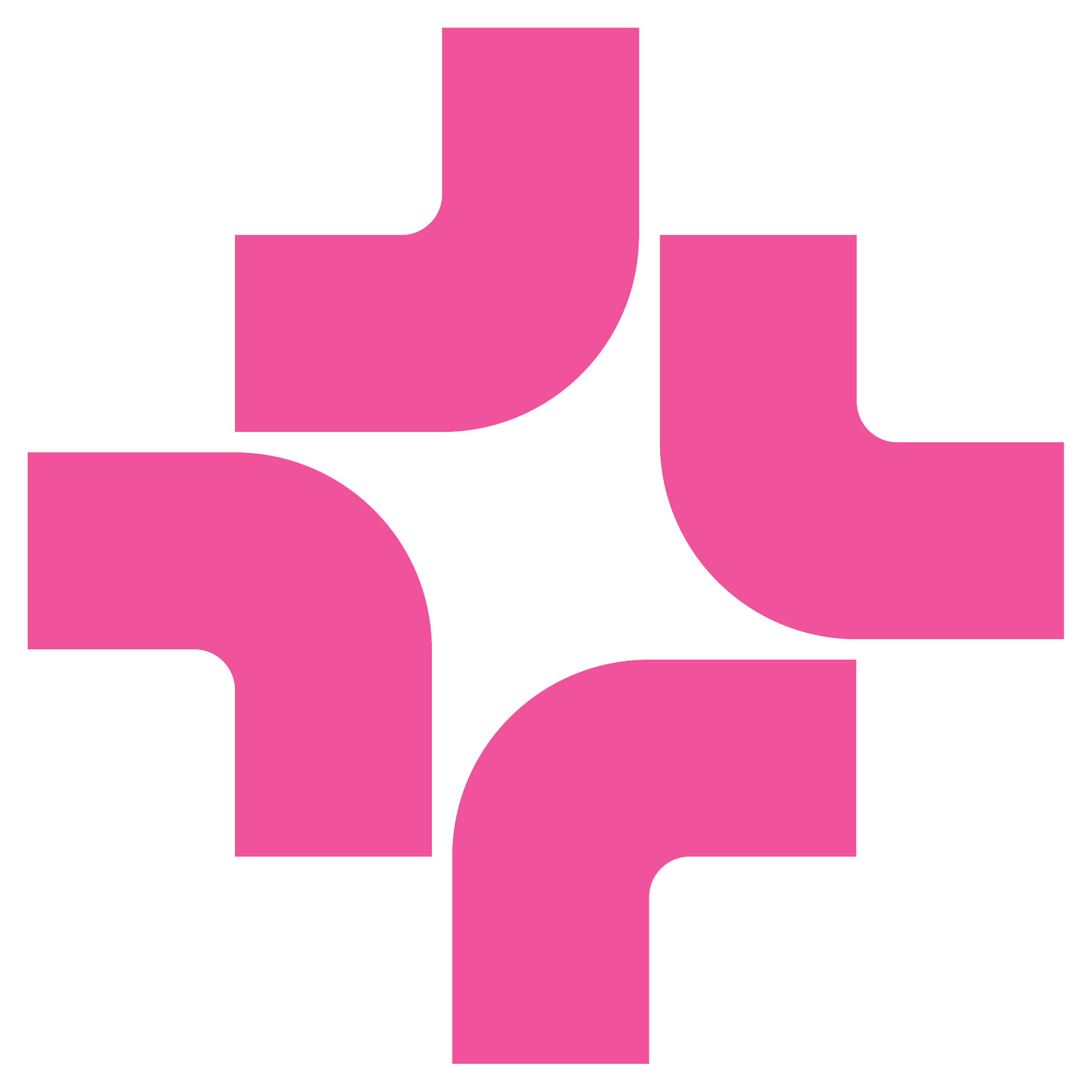}}\xspace Websites:]{\href{https://allenai.github.io/WildDet3D/}{\texttt{Technical-Website}} \quad \href{https://allenai.org/blog/wilddet3d}{\texttt{Blog}}}

\begin{document}

\maketitle

\section{Introduction}
\label{sec:introduction}

\begin{figure*}[t]
\centering
\includegraphics[width=\textwidth]{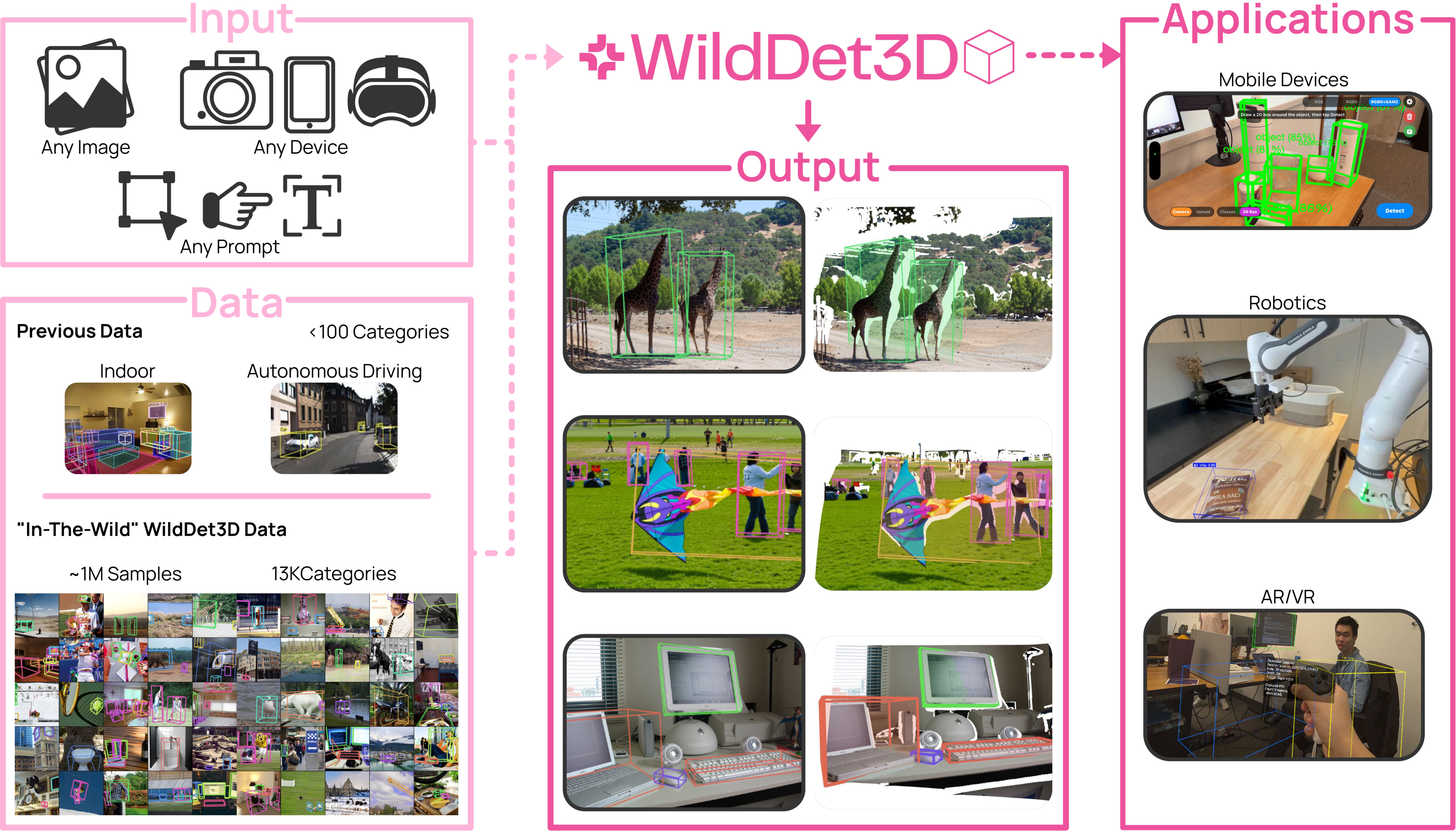}
\caption{\textbf{Overview of \model.} Given a single RGB image and an optional depth map, \model performs open-vocabulary monocular 3D object detection by accepting flexible prompt modalities---text queries, 2D point clicks, or 2D bounding boxes---and predicting full 3D bounding boxes for the specified objects. This unified framework enables interactive, open-world 3D perception across diverse scenes and thousands of object categories, supporting applications in mobile devices, robotics, and AR/VR. The model gracefully leverages additional geometric cues, \ie, depth when available. To train \model for broad generalization, we also curate \data, a large-scale in-the-wild dataset with approximately 1M human-verified samples spanning 13K categories.}
\label{fig:teaser}
\end{figure*}

Understanding objects in 3D is fundamental to spatial intelligence. An agent cannot reliably navigate, manipulate, or reason about the physical world by knowing what objects are alone; it must also understand where they are, how large they are, and how they are oriented in 3D space. This capability lies at the core of robotics, embodied AI, autonomous driving, and AR/VR use cases, where success depends on grounding perception in geometry rather than appearance alone. Despite rapid progress in open-vocabulary 2D object recognition driven by large-scale vision-language models, bringing the same flexibility to 3D remains difficult. Monocular 3D object detection---recovering the position, extent, and pose of objects from a single RGB image---is a core instance of this challenge, yet existing methods still lack the generality needed for open-world use.

What would it take to build a truly general-purpose monocular 3D detector? We argue that such a system must satisfy three requirements that are not well addressed by existing methods~\cite{brazil2023omni3dlargebenchmarkmodel,yang20253dmoodlifting2d3d,zhang2025detect3dwild,yao2025openvocabularymonocular3d}.
First, it should generalize in the wild, where object categories are long-tailed, open-ended, and frequently unseen during training.  
Second, it should support multiple prompt modalities. Different downstream applications naturally call for different ways of specifying a target object: a robot may issue a language command such as ``pick up the mug,'' an AR interface may let a user tap on a region of interest, and an upstream 2D detector may provide a bounding box to be lifted into 3D. A practical model should unify these interfaces---text, 2D points, and 2D boxes---within a single architecture rather than specialize to only one.
Third, real deployments may sometimes provide extra geometric cues, such as sparse LiDAR or partial depth, which should be leveraged to improve 3D localization when available. 
Prior work typically tackles only a subset of these requirements. Existing open-vocabulary methods~\cite{yang20253dmoodlifting2d3d,yao2025openvocabularymonocular3d} largely focus on text-based querying, whereas oracle-prompt methods~\cite{zhang2025detect3dwild,sam3dteam2025sam3d3dfyimages} assume fixed geometric inputs such as boxes; neither provides a flexible, unified framework for interactive open-world 3D perception, nor do they naturally accommodate additional depth signals at inference time.


\textbf{Addressing these challenges requires advances on both the model and data sides.} Handling flexible inputs demands a model that can unify text, points, and boxes within a single geometry-aware framework, while also incorporating
partial depth cues when available. At the same time, generalization in the wild depends on training data that captures broader vocabularies, greater visual diversity, and more realistic open-world settings. We present
contributions on both fronts, as highlighted in Figure~\ref{fig:teaser}.

\begin{figure}[!t]
  \centering
  \includegraphics[width=\linewidth]{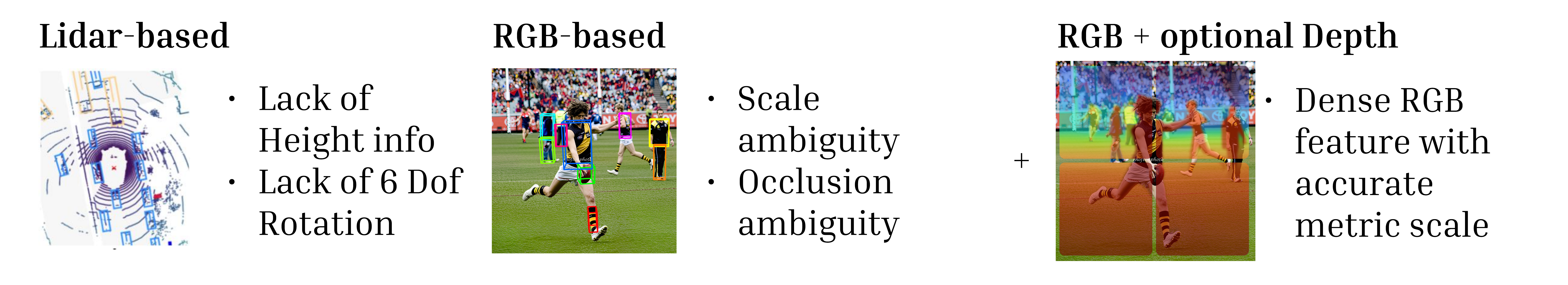}
  \caption{\textbf{Input modality comparison for generalized 3D detection.} LiDAR point clouds lack reliable height information and full 6-DoF rotation cues.  RGB images provide dense appearance features but suffer from inherent scale and occlusion ambiguity. By combining RGB with \emph{optional} depth, our approach retains the rich visual semantics needed for open-vocabulary recognition while reducing metric scale ambiguity when geometric signals are available.}
  \label{fig:modality}
\end{figure}

On the model front, we introduce \textbf{\model}, a state-of-the-art open model for open-vocabulary monocular 3D detection.
\model handles multiple prompt modalities---text, 2D points, and 2D boxes---within a single geometry-aware architecture, making it suitable for flexible and interactive open-world 3D perception. By combining strong open-vocabulary visual recognition with monocular geometry estimation, it predicts 3D bounding boxes from a single image, while also leveraging additional geometric cues such as partial depth when available. A key design question for a generalized 3D detector is the choice of input modality (Figure~\ref{fig:modality}). LiDAR-based methods provide direct geometric measurements but produce sparse point clouds that lack reliable height information and full 6-DoF rotation cues, limiting their applicability to categories with well-defined upright
priors. Pure RGB approaches offer dense visual features suitable for open-vocabulary recognition, yet they face inherent scale ambiguity---a small nearby object is indistinguishable from a large distant one---and occlusion ambiguity when objects overlap. We argue that \emph{RGB with optional depth} strikes the best balance: dense
appearance features support open-vocabulary, open-world recognition across arbitrary categories, while depth---when available from a LiDAR, stereo pair, or sensor---resolves metric scale without sacrificing visual richness. Crucially, by making depth \emph{optional}, the model degrades gracefully to monocular mode rather than failing when geometric signals are absent (Section~\ref{sec:method}).

On the data front, we introduce \textbf{\data}, a large-scale in-the-wild dataset for 3D detection that complements the model's generalization ability. We build it by applying existing models and methods to generate candidate
3D boxes for 2D annotations drawn from diverse 2D detection datasets~\cite{lin2015microsoftcococommonobjects,9009553,gupta2019lvis,wang2023v3detvastvocabularyvisual},
and then asking human annotators to select the best qualified boxes, if any, from these candidates. This process produces a curated dataset of over 1M images covering 13.5K categories in diverse real-world scenes, substantially expanding vocabulary coverage and scene diversity for monocular 3D detection. The resulting human-verified supervision enables \model to generalize well beyond standard benchmark settings (Section~\ref{sec:data}).

Empirically, \model delivers strong performance across open-world in-the-wild evaluation, standard benchmarks, and zero-shot transfer. On \bench, our in-the-wild benchmark spanning 700+ open-vocabulary categories, \model achieves 22.6~AP$_\text{3D}$ with text prompts and 24.8~AP$_\text{3D}$ with box prompts, far exceeding prior methods (2.3~AP for 3D-MOOD). When ground-truth depth is provided, performance reaches 41.6~AP (text) and 47.2~AP (box). On Omni3D~\cite{brazil2023omni3dlargebenchmarkmodel}, it surpasses prior methods in both text-prompt and box-prompt (oracle) settings, achieving 34.2 and 36.4 AP$_\text{3D}$, respectively, while training for only 12 epochs, compared with 80--120 epochs for competing approaches. The model also generalizes effectively across datasets: trained on Omni3D and evaluated zero-shot, it reaches 40.3 ODS on Argoverse~2~\cite{wilson2023argoverse2generationdatasets} and 48.9 ODS on ScanNet~\cite{dai2017scannet}, with particularly large improvements on novel categories unseen during training. Moreover, when extra geometric cues such as sparse or ground-truth depth are available at inference time, the gains are substantial, highlighting the model’s ability to flexibly benefit from richer 3D information (Section~\ref{sec:experiments}).

Finally, we demonstrate the versatility of \model across a range of real-world deployment scenarios, including an interactive web demo, on-device iPhone inference, AR/VR integration with Meta Quest and Meta Glasses, vision-language model integration for spatial reasoning, and robotic manipulation. These applications highlight that \model serves as a general-purpose 3D perception module that can be deployed across platforms and paired with diverse upstream systems in a plug-and-play fashion (Section~\ref{sec:application}).

\section{WildDet3D}
\label{sec:method}

\begin{figure*}[t]
\centering
\includegraphics[width=\linewidth]{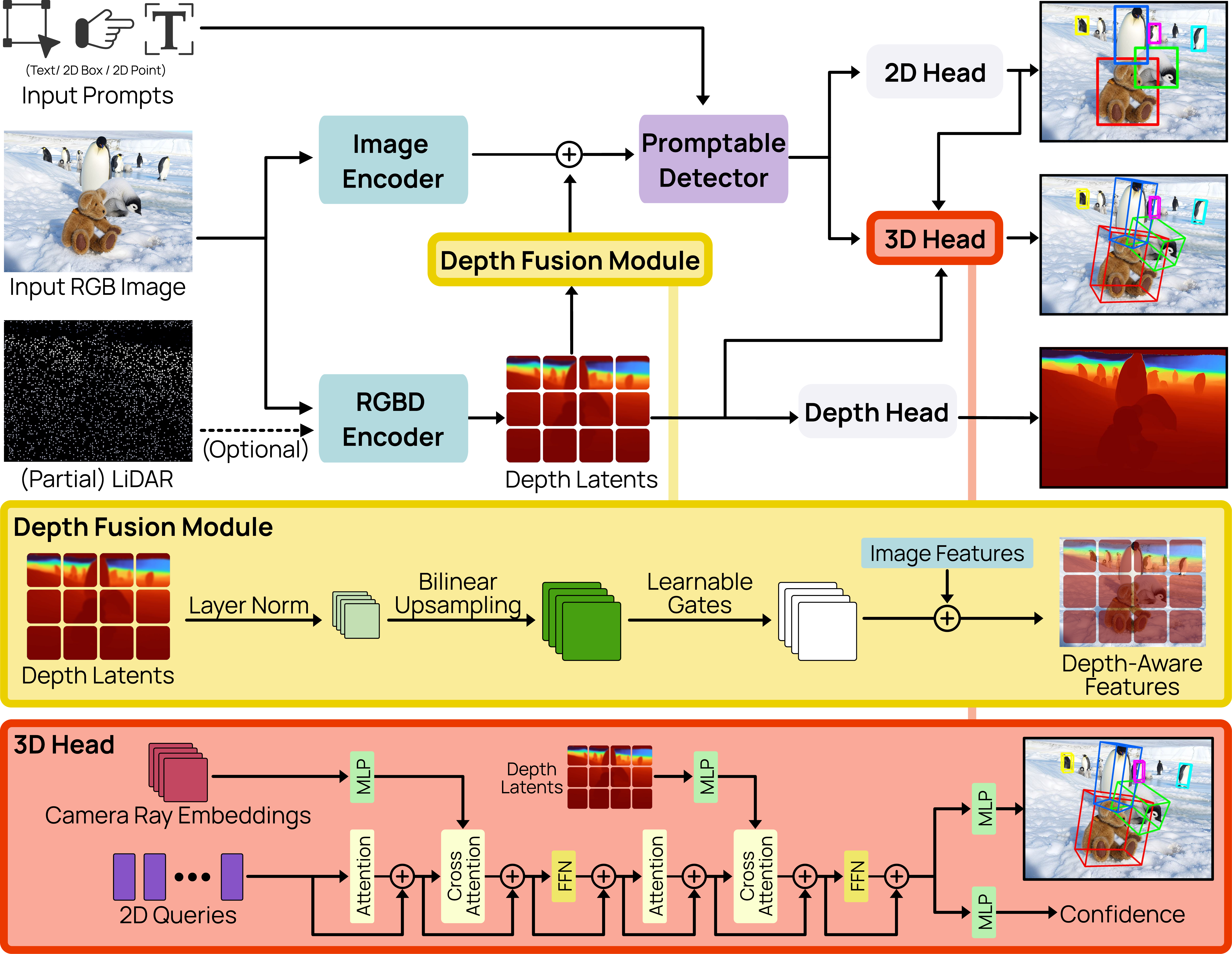}
\caption{\textbf{Overview of \model.}
Given an RGB image and optional depth input, {\color{ai2blue} dual-vision encoders (Image + RGBD)} extract visual features and depth latents in parallel. The {\color{yellow}depth fusion module} processes the depth latents generated by the RGBD encoder and combines it with image features from the image encoder via element-wise addition to produce enriched visual queries, which are then integrated with diverse input prompts via the {\color{ai2purple}promptable detector}. The resulting outputs are passed through cascaded 2D and {\color{red}3D Heads} for open-vocabulary object detection, while the depth latents are separately decoded for depth estimation. Although our primary focus is 3D object detection, the auxiliary {\color{gray}2D detection and depth heads} provide complementary supervision that improves overall performance.
}
\label{fig:pipeline}
\end{figure*}

Given a single RGB image $\mathbf{I}\!\in\!\mathbb{R}^{3\times H\times W}$, optional camera intrinsics $\mathbf{K}\!\in\!\mathbb{R}^{3\times 3}$, optional partial or full depth map $\mathbf{D}\!\in\!\mathbb{R}^{H\times W}$ (\eg, from LiDAR, stereo, or ToF sensors), and a user-specified prompt $\mathcal{P}$, \model predicts a set of 3D bounding boxes $\{\mathbf{B}_i\}_{i=1}^{N}$ for all target objects in the scene. Each 3D box $\mathbf{B}_i = (\mathbf{c}_i, \mathbf{d}_i, \mathbf{R}_i, s_i)$ consists of a 3D center $\mathbf{c}_i \in \mathbb{R}^3$ in metric space (meters), physical dimensions $\mathbf{d}_i = (w, h, l) \in \mathbb{R}^3_{+}$ in meters, an orientation $\mathbf{R}_i \in \text{SO}(3)$ with unambiguous rotation normalization, and a confidence score $s_i \in [0, 1]$. When intrinsics or depth are not provided, the model falls back to its built-in estimation modules.

An overview of the architecture of \model is shown in Figure~\ref{fig:pipeline}. As illustrated, \model comprises three main components: 
(1) to accommodate optional geometric signals (\eg, input depth available at inference time), we introduce dual-vision encoders ({\color{ai2blue} blue blocks}) and a depth fusion module ({\color{yellow} yellow block}) that produce geometry-aware depth latents (Section~\ref{subsec:encoder});
(2) to unify different forms of input prompts within a single architecture, we develop a promptable detector ({\color{ai2purple} purple block}) that takes the fused vision features along with input prompts to produce unified query representations for detection heads (Section~\ref{subsec:promptable_detector});
(3) for 3D bounding box prediction, we propose a 3D detection head with unambiguous rotation normalization ({\color{red} red block}) that aggregates multi-source information spanning depth, 2D spatial, and semantic features (Section~\ref{subsec:3d_head}). We further introduce auxiliary 2D detection and depth estimation heads ({\color{gray} gray blocks}) that substantially boost 3D performance while enabling broader downstream applications (Section~\ref{subsec:loss}) .

\subsection{Dual-vision encoder}
\label{subsec:encoder}
Accurate 3D detection from monocular images poses two intertwined challenges: the detection backbone must extract rich semantic features for recognition and localization, while the system simultaneously requires metrically grounded depth and camera-aware representations to reason in 3D.
Naïvely coupling these two objectives inside a single encoder forces a trade-off---optimizing for depth can degrade detection features and vice versa---and tightly binds the architecture to one particular depth model.

We address this with a \emph{dual-encoder} design comprising three components.
First, an \textbf{image encoder} provides high-resolution, multi-scale semantic features.
Second, an \textbf{RGBD encoder} operates as a pluggable geometry backend: it ingests the same image plus an optional partial or full depth map from a pretrained RGBD backbone and produces depth latents through a multi-level convolutional neck.
Third, a \textbf{depth fusion module} merges the geometric cues from the RGBD encoder into the semantic feature maps of the image encoder, supplying the downstream detection head with a unified representation that is both semantically rich and metrically informed.
By decoupling semantics from geometry at the encoder level and reuniting them through a dedicated fusion stage, the architecture remains modular---different depth models can be integrated without modifying the core detection pipeline.

\paragraph{Image encoder.}
The image encoder is a ViT-H~\cite{dosovitskiy2021imageworth16x16words} with a SimpleFPN neck, initialized from a segmentation-pretrained checkpoint that provides strong dense prediction features. Given an input image resized to $H\!\times\!W$ pixels with patch size $p$, the ViT generates $\frac{H}{p}\!\times\!\frac{W}{p}$ spatial tokens. The SimpleFPN projects these to 256-channel feature maps that are fed to the downstream detector.
During training, the first 28 of 32 ViT blocks are frozen, with only the last 4 blocks fine-tuned.
We adopt the image encoder architecture and weights from SAM~3~\cite{carion2025sam3segmentconcepts}, which provides high-quality dense features for detection and segmentation.

\paragraph{RGBD encoder.}
The RGBD encoder is built on a DINOv2 ViT-L/14~\cite{oquab2024dinov2learningrobustvisual} that accepts 4-channel RGBD input at $686\!\times\!686$ resolution ($49\!\times\!49$ tokens), where the depth channel is optional---when no external depth is available, a zero-filled depth channel is used and RGBD encoder will generate depth feature solely based on RGB. The encoder features are passed through a ConvStack neck that produces a 5-level feature pyramid, from which we extract depth latents $\mathbf{Z}_d \in \mathbb{R}^{C_d\times 49\times 49}$ with $C_d\!=\!256$ via average pooling.
During training, the first 21 of 24 DINOv2 blocks are frozen, with the last 3 blocks fine-tuned. To support optional depth input, training uses a stochastic strategy: 70\% monocular (zero depth), 20\% patch-masked depth, and 10\% full depth copy-through.
The RGBD encoder architecture and weights are adopted from LingBot-Depth~\cite{lingbotdepth2026}, a model pretrained for metric depth estimation on large-scale RGBD data.

\emph{We deliberately use different backbones for the image and RGBD encoders}: the image encoder is pretrained for segmentation, providing detection-oriented features, while the RGBD encoder is pretrained on depth completion, producing geometric features suited for metric depth estimation.

\paragraph{Depth fusion module.}
The depth fusion module (yellow block in Figure~\ref{fig:pipeline}) injects depth latents into the image encoder's feature maps before they enter the transformer encoder, following a ControlNet-style~\cite{zhang2023addingconditionalcontroltexttoimage} residual design.
Given visual features $\mathbf{V} \in \mathbb{R}^{C \times H_v \times W_v}$ from the SimpleFPN and depth latents $\mathbf{Z}_d$, the module computes:
\begin{equation}
  \mathbf{V}' = \mathbf{V} + \text{Conv}_{1\times1}\!\big(\text{LN}(\mathbf{Z}_d^{\uparrow})\big),
  \label{eq:depth_fusion}
\end{equation}
where $\mathbf{Z}_d^{\uparrow}$ denotes depth latents bilinearly interpolated to match the visual feature resolution, $\text{LN}$ is LayerNorm that normalizes the depth latents to unit scale, and $\text{Conv}_{1\times1}$ is a $1\!\times\!1$ convolution projecting from depth dimension to visual dimension.
The convolution is \emph{zero-initialized}, so at training start $\mathbf{V}' = \mathbf{V}$ (identity), and the depth contribution is gradually learned without disrupting pretrained visual features.
Notably, only the depth branch passes through trainable layers; the visual features are added as-is, preserving the pretrained feature distribution.

\subsection{Promptable detector}
\label{subsec:promptable_detector}
The promptable detector conditions metric-depth-aware visual features on user-supplied prompts to produce per-object predictions. It accepts four complementary prompt types:
\begin{itemize}[nosep]
  \item \textbf{Text prompt.} A category name (\eg, ``car''),
        selecting all instances of that category.
  \item \textbf{Point prompt.} One or more 2D pixel coordinates $(u,v)$, each labeled as
        positive (on the object) or negative (background), selecting the single object at that location \cite{clark2025molmopoint,molmov1,yuan2024robopoint}.
  \item \textbf{Box prompt.} A 2D bounding box
        $(x_1,y_1,x_2,y_2)$, selecting the single object within the specified region.
  \item \textbf{Exemplar prompt.} A 2D bounding box used as a visual exemplar,
        detecting all visually similar objects in the scene.
\end{itemize}
During training, all four prompt types are sampled jointly to ensure balanced learning across modalities.

\paragraph{Prompt encoding.}
We adopt SAM3's prompt encoding~\cite{carion2025sam3segmentconcepts} as described below.
\emph{Text prompts} are tokenized with a CLIP-style~\cite{radford2021learningtransferablevisualmodels} BPE tokenizer and encoded by a 24-layer causal text Transformer (width 1024, 16 heads), then linearly projected to $d\!=\!256$.
\emph{Box and point prompts} are encoded by a geometry encoder that sums three complementary representations: (1) a direct linear projection of the coordinates, (2) ROI-aligned features pooled from the image backbone (for boxes) or grid-sampled features (for points), and (3) sinusoidal positional encoding. A learnable positive/negative label embedding is added, and the result is refined by a 3-layer Transformer with cross-attention to image features.
\emph{Exemplar prompts} reuse the same box encoding pipeline but are differentiated by a special text token (``visual'') and a multi-target matching strategy that assigns all instances of the same category as ground truth.
The encoded text and geometry tokens are concatenated into a single prompt sequence, which serves as cross-attention memory in both the encoder and decoder stages.

\paragraph{Per-prompt batching.}
Rather than constructing the training batch at the per-image level, we batch at the \emph{per-prompt} level. For example, every unique text category yields a separate batch entry that aggregates all images containing that category.  This strategy enables fine-grained multi-instance supervision and allows the model to handle an arbitrary number of categories per image without padding or truncation.

\subsection{Deeply-supervised 3D detection head}
\label{subsec:3d_head}
The 3D detection head (red block in Figure~\ref{fig:pipeline}) lifts the 2D query features produced by the promptable detector into 3D bounding-box predictions.  
It comprises $L$ Transformer decoder layers, each of which outputs its own set of 3D predictions; the training losses, which will be detailed in Section~\ref{subsec:loss}, are applied at every layer with equal weights. This \emph{deep-supervision} strategy encourages even the earliest layers to develop 3D localization capability, yielding faster convergence and more robust intermediate representations. The remainder of this section details the individual components of the head.

\paragraph{Multi-source information aggregation.}
For each decoder layer $l \in \{1,\ldots,L\}$, the hidden states
$\mathbf{H}^l \in \mathbb{R}^{S\times d}$ (where $d\!=\!256$) are
sequentially enriched with camera geometry and depth information
through two dedicated cross-attention modules.
First, a \emph{camera prompt} branch incorporates spatial ray features.
Given camera intrinsics $\mathbf{K}$, we generate per-pixel ray directions $\mathbf{r}_{i,j} = \mathbf{K}^{-1}[u, v, 1]^\top$ and encode them using 8th-order real spherical harmonics:
\begin{equation}
  \phi(\mathbf{r}) = \text{RSH}_8\!\left(\frac{\mathbf{r}}{\|\mathbf{r}\|}\right)
  \in \mathbb{R}^{81},
  \label{eq:rsh}
\end{equation}
where $\text{RSH}_8$ denotes the 8th-order spherical harmonic basis functions.
The ray features are then fused via cross-attention:
\begin{equation}
  \tilde{\mathbf{H}}^l =
  \text{FFN}\;\!\Big(
    \text{CrossAttn}\;\!\big(
      \text{SelfAttn}(\mathbf{H}^l),\;
      f_r(\phi(\mathbf{r}))
    \big)
  \Big),
  \label{eq:camera_prompt}
\end{equation}
where $f_r\!: \mathbb{R}^{81}\!\to\!\mathbb{R}^d$ projects the
spherical harmonic ray features.
Then, a \emph{depth prompt} branch fuses depth latents:
\begin{equation}
  \hat{\mathbf{H}}^l =
  \text{FFN}\;\!\Big(
    \text{CrossAttn}\;\!\big(
      \text{SelfAttn}(\tilde{\mathbf{H}}^l),\;
      f_d(\mathbf{Z}_d)
    \big)
  \Big),
  \label{eq:depth_prompt}
\end{equation}
where $f_d\!: \mathbb{R}^{C_d}\!\to\!\mathbb{R}^d$ is a learned
projection that maps depth latents to the query embedding space.
Both cross-attention modules use a single attention head, and each
decoder layer has its own independent set of projection parameters.

\paragraph{3D box parameterization.}
The fused query features are passed through a two-layer MLP to predict a 12-dimensional 3D box encoding:
\begin{equation}
  \mathbf{p}_\text{3d} = \big(
    \underbrace{\Delta c_x, \Delta c_y}_{\text{center offset}},\;
    \underbrace{\hat{d}}_{\text{log depth}},\;
    \underbrace{\hat{w}, \hat{h}, \hat{l}}_{\text{log dims}},\;
    \underbrace{r_1, \ldots, r_6}_{\text{rotation}}
  \big).
  \label{eq:3d_param}
\end{equation}

\noindent
The components are defined as:

\begin{itemize}[nosep]
\item \textbf{Center offset}
  $(\Delta c_x, \Delta c_y)$:
  the displacement between the 2D projection of the 3D center and the
  2D box center, normalized by a scale factor $s_c\!=\!10$.

\item \textbf{Log-depth}
  $\hat{d} = s_d \cdot \log(d)$:
  the logarithm of the metric depth, scaled by $s_d\!=\!2.0$.

\item \textbf{Log-dimensions}
  $(\hat{w}, \hat{h}, \hat{l}) = s_\text{dim} \cdot \log(w, h, l)$:
  the logarithm of physical dimensions in meters, scaled by
  $s_\text{dim}\!=\!2.0$.

\item \textbf{6D rotation}
  $(r_1,\ldots,r_6)$:
  the first two rows of the $3\!\times\!3$ rotation matrix, following
  the continuous 6D representation~\cite{zhou2020continuityrotationrepresentationsneural}, from
  which the full rotation matrix is recovered via Gram--Schmidt
  orthogonalization.
\end{itemize}

\paragraph{Unambiguous rotation normalization.}
Oriented 3D bounding boxes have an inherent rotation ambiguity: a box with dimensions $(w, h, l)$ rotated by yaw $\theta$ is geometrically identical to one with swapped dimensions $(l, h, w)$ rotated by $\theta + 90^\circ$, and similarly a $180^\circ$ yaw flip yields the same box for symmetric objects.
Without normalization, the model must learn multiple equivalent representations for the same 3D box, which makes training harder.

We resolve this with a two-step unambiguous rotation normalization applied to the ground-truth rotation and dimensions before loss computation:
(1) \emph{Dimension ordering}: if $w > l$, swap $(w, l)$ and rotate by $R_y(90^\circ)$ so that $w \leq l$ always holds.
(2) \emph{Yaw folding}: fold the yaw angle into $[0, \pi)$ by applying $R_y(180^\circ)$ when yaw $< 0$ or yaw $\geq \pi$.
Together, these two steps reduce a 4-fold rotation--dimension ambiguity to a unique unambiguous form, yielding a one-to-one mapping between box geometry and the regression target. The same normalization is applied to predictions at inference time before evaluation.

\noindent
At inference, the 3D center is recovered by adding the predicted
offset to the 2D box center, then back-projecting through
$\mathbf{K}^{-1}$ at the predicted depth $d = \exp(\hat{d}/s_d)$.

\paragraph{3D confidence prediction.}
In addition to the box regression branch, we introduce a parallel
confidence branch---a two-layer MLP---that
predicts a scalar 3D detection quality score $s_\text{3D} \in [0,1]$.
During training, the soft target is defined as:
\begin{equation}
  q^* = \beta \cdot q_{\text{depth}} + (1-\beta) \cdot
  \text{IoU}_\text{3D},
  \label{eq:3d_conf}
\end{equation}
where $q_\text{depth} = \exp(-|\log \hat{d} - \log d^*|)$ measures depth prediction quality as a symmetric ratio bounded in $[0,1]$,
$\text{IoU}_\text{3D}$ is the 3D box IoU with the matched ground
truth, and $\beta\!=\!0.7$ to emphasize depth accuracy, which is the primary bottleneck in monocular 3D detection.
At inference, the final detection score combines the 2D objectness score $s_\text{2D}$ (from the IoU-aware classification head) and the 3D confidence $s_\text{3D}$:
\begin{equation}
  s = s_\text{2D} + \alpha \cdot s_\text{3D},
  \label{eq:final_score}
\end{equation}
with $\alpha\!=\!0.5$.
This additive formulation allows the 3D confidence to re-rank detections that have similar 2D scores but differ in geometric quality, while keeping $s_\text{2D}$ as the dominant term so that high-confidence 2D detections are not suppressed by uncertain 3D estimates.

\subsection{Multi-task learning}
\label{subsec:loss}
During training, each category in an image produces two branches of queries:
(1)~a \emph{multi-target} query---sampled as 50\% text-only and 50\% exemplar box (with optional category label)---that is supervised against all instances of that category, and
(2)~a \emph{single-target} geometric query (box or point prompt with optional category label) that is directly assigned to one selected instance.
This dual-branch design ensures all prompt modalities receive supervision simultaneously.

Independently of the query branch, we employ one-to-many (O2M) matching~\cite{carion2025sam3segmentconcepts}: each ground-truth object is paired with its top-$k$ scoring predictions ($k{=}4$), providing denser supervision that accelerates convergence.

The overall training loss aggregates 3D regression, 3D confidence losses, geometry estimation loss, and 2D detection loss:
\begin{equation}
  \mathcal{L} =
    \underbrace{
    \mathcal{L}_\text{3D} +
    \mathcal{L}_\text{conf}}_{\text{3D detection losses}} +
    \underbrace{
    \mathcal{L}_\text{geom} +
    \mathcal{L}_\text{2D}}_{\text{auxiliary losses}}.
  \label{eq:total_loss}
\end{equation}

\paragraph{3D regression loss $\mathcal{L}_\text{3D}$.}
For each matched prediction--target pair, we compute an L1 loss on the
encoded 3D parameters (Eq.~\ref{eq:3d_param}):
\begin{equation}
  \mathcal{L}_\text{3D} = \frac{1}{N_\text{pos}}\!\sum_{i \in \mathcal{M}}
  \sum_{k} w_k \left| p_k^{(i)} - p_k^{*(i)} \right|,
  \label{eq:3d_loss}
\end{equation}
where $\mathcal{M}$ is the set of matched indices, $p_k$ and $p_k^*$
are the $k$-th components of the predicted and target encodings, and
$w_k$ are per-component validity weights (set to zero when depth or
dimensions are unavailable in the annotation).

\paragraph{3D confidence loss $\mathcal{L}_\text{conf}$.}
The confidence branch is trained with an IoU-aware focal BCE loss.
For each matched prediction with raw logit $c_i$, we construct an adaptive soft target:
\begin{equation}
  t_i = \sigma(c_i)^{\alpha} \cdot {q_i^*}^{1-\alpha},
  \label{eq:conf_target}
\end{equation}
where $\sigma(\cdot)$ is the sigmoid function, $q_i^*$ is the regression quality (Eq.~\ref{eq:3d_conf}), and $\alpha\!=\!0.25$.
The total confidence loss combines a positive term over matched queries and a focal-weighted negative term over unmatched queries:
\begin{equation}
  \mathcal{L}_\text{conf} =
  \frac{1}{N_\text{pos}}\sum_{i \in \mathcal{M}}
    w_+ \cdot \text{BCE}(c_i, t_i)
  + \frac{1}{N_\text{neg}}\sum_{j \notin \mathcal{M}}
    \sigma(c_j)^{\gamma} \cdot \text{BCE}(c_j, 0),
  \label{eq:conf_loss}
\end{equation}
where $w_+\!=\!5$ is a positive sample weight and $\gamma\!=\!2$ is the focal exponent that down-weights easy negatives.

\paragraph{Auxiliary geometry loss $\mathcal{L}_\text{geom}$.}
The geometry backend loss aggregates below losses on the predicted depth map and camera intrinsics:
\begin{itemize}[nosep]
\item \textbf{L1 metric depth} loss between predicted and ground-truth depth at valid pixels;
\item \textbf{scale-invariant logarithmic depth} loss~\cite{eigen2014depthmappredictionsingle} between predicted and ground-truth depth at valid pixels;
\item \textbf{confidence mask binary cross-entropy} loss that supervises a per-pixel depth validity prediction;
\item \textbf{affine-invariant point-map} losses (global alignment, multi-scale local alignment, and edge-aware losses)~\cite{wang2025moge2accuratemonoculargeometry} computed on
  back-projected 3D point maps for geometric consistency;
\item \textbf{camera ray directions L2} loss that supervises the predicted intrinsics against ground-truth camera parameters.
\end{itemize}
For completeness, the detailed formulations are provided in Appendix.

\paragraph{Auxiliary 2D detection loss $\mathcal{L}_\text{2D}$.}
The 2D detection loss aggregates below losses:
\begin{itemize}[nosep]
\item \textbf{IoU-aware binary cross-entropy} loss~\cite{carion2025sam3segmentconcepts} for box classification, where the soft target is
the 2D IoU between the predicted and ground-truth boxes;
\item \textbf{box regression} combines L1 loss on the center-size representation and generalized IoU loss~\cite{rezatofighi2019generalizedintersectionunionmetric} for bounding boxes in pixel-space.  
\item \textbf{per-category presence} loss supervises whether a queried category exists in the image.  
\item \textbf{one-to-many matching} loss: each ground-truth object is paired with its top-$k$ ($k{=}4$) scoring predictions (see above), providing denser gradient signals for both 2D and 3D heads.
\end{itemize}
For completeness, the detailed formulations are provided in Appendix.

\paragraph{Ignore-region suppression.}
A fundamental challenge in monocular 3D detection is \emph{non-exhaustive annotation: not every visible object has a valid 3D ground truth}.  
For example, in both Omni3D~\cite{brazil2023omni3dlargebenchmarkmodel} and \data, objects with invalid 3D measurements, heavy truncation, severe occlusion, or placement behind the camera are annotated as \texttt{IGNORE}---they retain their 2D bounding boxes but are excluded from the set of positive 3D targets.
We address this challenge in a consistent way for both evaluation and training.
During \textbf{evaluation}, a prediction that matches an ignored ground-truth box is treated as \emph{neutral}: it counts as neither a true positive nor a false positive, and ignored ground truths do not contribute to the false-negative count.
During \textbf{training}, we adopt the \emph{ignore-region suppression} strategy widely used in 2D methods~\cite{lin2015microsoftcococommonobjects,Geiger2012CVPR,cordts2016cityscapes}. Concretely, we suppress the negative classification loss for any prediction whose 2D~IoU with an ignore-annotated box exceeds $0.5$ (2D IoU is used because ignore-annotated objects lack valid 3D ground truth).
This ensures the training objective is consistent with the evaluation protocol, allowing the model to confidently detect objects regardless of whether their 3D annotations are available.


\section{WildDet3D-Data}
\label{sec:data}

Existing 3D detection datasets such as Omni3D~\cite{brazil2023omni3dlargebenchmarkmodel} provide high-quality annotations but suffer from two key limitations: (1)~\textbf{limited scale}, typically covering fewer than 100 categories, and (2)~\textbf{narrow domain coverage}, focusing on settings such as autonomous driving or indoor scenes. Scaling 3D detection to the open world is fundamentally more challenging than its 2D counterpart: unlike 2D bounding boxes, 3D annotations require metric depth and calibrated camera intrinsics, both of which are costly to obtain at scale.

\begin{figure}[t]
  \centering
  \includegraphics[width=\textwidth]{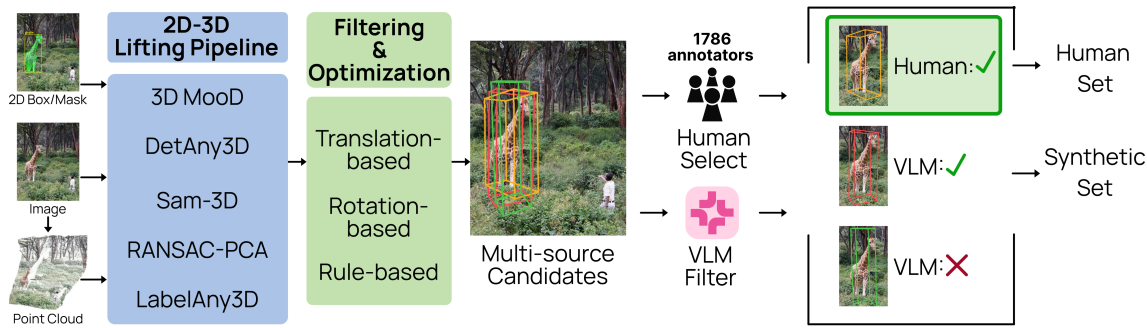}
  \caption{\textbf{Overview of the \data pipeline.}
  Given an image with 2D boxes or masks and a depth-derived point cloud, five complementary models generate candidate 3D bounding boxes, shown in different colors. These candidates are refined through translation and rotation optimization, followed by rule-based filtering. The filtered candidates then enter two parallel selection branches: a VLM filtering branch, which scores each candidate on six perceptual criteria and retains those whose scores exceed a threshold, and a human annotation branch, in which annotators select the best candidate and assess its quality.}
  \label{fig:data_pipeline}
\end{figure}

We introduce \data, a large-scale dataset for open-vocabulary 3D detection in the wild. 
Our dataset covers over \textbf{1M images} across \textbf{22 scene categories} (shown in Figure~\ref{fig:scene_distribution}), with \textbf{3.7M valid 3D annotations}, and \textbf{13.5K object categories}---a \textbf{138$\times$} increase in category coverage over Omni3D.
To collect this dataset, we develop a three-stage pipeline:
(1)~multiple complementary models generate candidate 3D boxes for each existing 2D annotation (Section~\ref{subsec:candidate_gen}),
(2)~rule-based geometric and semantic filters remove implausible candidates (Section~\ref{subsec:quality_control}),
and (3)~human annotators or VLM-based selectors choose the best candidate and rate its quality (Section~\ref{subsec:annotation}).
An overview of the pipeline is shown in Figure~\ref{fig:data_pipeline}. Selected dataset samples are visualized in Figure~\ref{fig:qualitative_data}. More examples are available in Appendix~\ref{appendix:qualitative_dataset}.

\begin{figure*}[t]
  \centering
  \includegraphics[width=\textwidth]{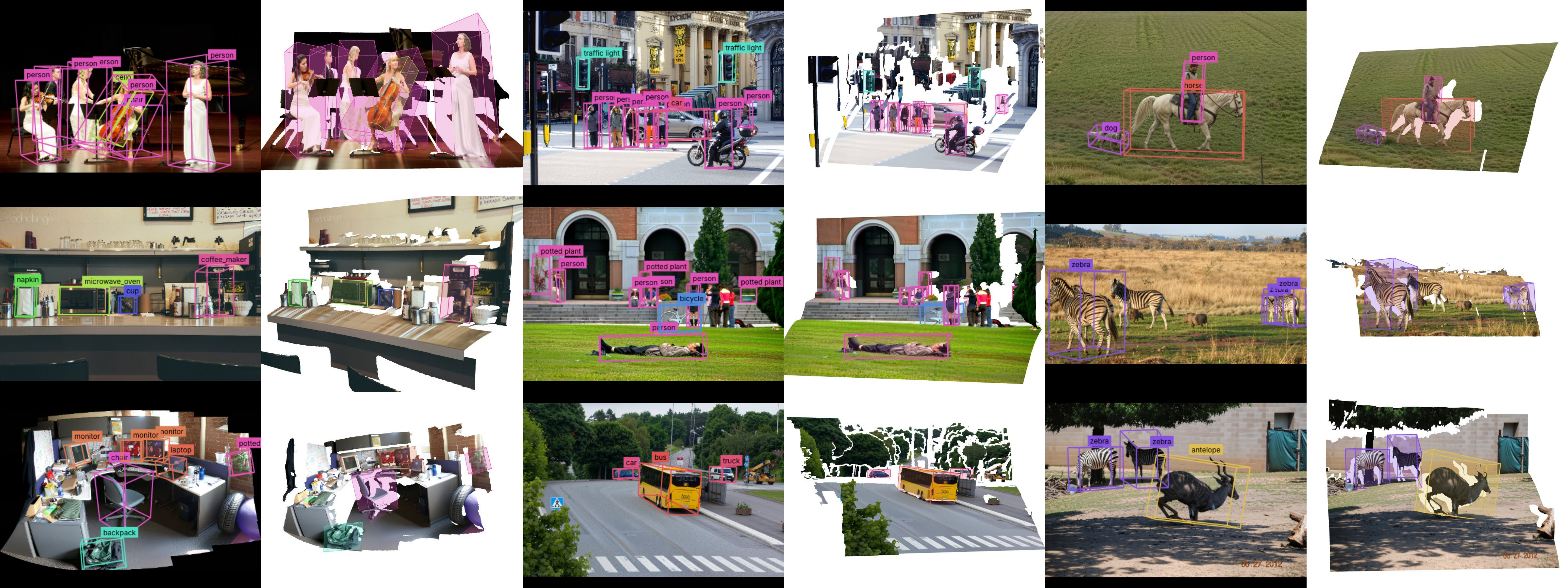}
  \caption{\textbf{Qualitative examples from \data.}
  Each pair shows 3D bounding box annotations overlaid on the input image with category labels (left) and the corresponding 3D bounding boxes rendered in the reconstructed point cloud (right).
  The dataset covers diverse settings including indoor scenes (desks, kitchens), outdoor environments (floating markets, streets), and animals in the wild, spanning a wide range of object categories, scales, and spatial layouts.}
  \label{fig:qualitative_data}
\end{figure*}

\subsection{Candidates generation}
\label{subsec:candidate_gen}

Each image undergoes a multi-step processing pipeline to produce
candidate 3D bounding boxes for existing 2D annotation.

\paragraph{Data sources.}
We draw 2D bounding box annotations from four large-scale detection
datasets:
\textbf{COCO}~\cite{lin2015microsoftcococommonobjects} (118K train, 5K val images),
\textbf{LVIS}~\cite{gupta2019lvis} (using COCO images with long-tail
annotations over 1,200+ categories),
\textbf{Objects365}~\cite{9009553} (609K train, 30K val images,
365~categories), and
\textbf{V3Det}~\cite{wang2023v3detvastvocabularyvisual} (183K train, 30K val images,
13K+ fine-grained categories).
Together these provide dense, high-quality 2D annotations with broad
category and scene diversity, forming the foundation for 3D lifting.

\paragraph{Monocular depth and camera estimation.}
We first apply 4$\times$ image super-resolution~\cite{yue2025arbitrarystepsimagesuperresolutiondiffusion} to
increase spatial detail for downstream point cloud generation.
MoGe-2~\cite{wang2025moge2accuratemonoculargeometry} then produces a metric depth map at
1024-long-edge resolution, while PerspectiveFields~\cite{jin2023perspectivefieldssingleimage}
and WildCamera~\cite{zhu2023tame} estimate camera pose (roll/pitch)
and intrinsics ($f_x, f_y, c_x, c_y$), respectively.
The depth map is reprojected into a 3D point cloud using the estimated
camera parameters.

\paragraph{Multi-model candidate generation.}
Five complementary methods generate candidate 3D boxes per
2D annotation, each capturing different geometric cues:
\begin{itemize}[nosep]
\item \textbf{3D-MOOD}~\cite{yang20253dmoodlifting2d3d}: Runs open-vocabulary text-based
  detection and matches predictions to ground-truth 2D boxes via IoU.
\item \textbf{DetAny3D}~\cite{zhang2025detect3dwild}: Directly regresses a 3D box
  from each 2D box using dense feature extraction.
\item \textbf{SAM-3D}~\cite{sam3dteam2025sam3d3dfyimages}: Reconstructs a 3D mesh from the
  object mask and depth-derived point map, then extracts an oriented
  bounding box from the mesh vertices.
\item \textbf{RANSAC-PCA}: A purely geometric method that extracts
  object points via masks, applies statistical outlier removal and
  HDBSCAN clustering, then fits an oriented box using RANSAC rectangle
  fitting with PCA-based gravity alignment.
\item \textbf{LabelAny3D}~\cite{yao2026labelany3dlabelobject3d}: Single-image 3D
  reconstruction that lifts 2D crops to 3D meshes and aligns them
  to the scene depth.
\end{itemize}

\paragraph{3D box optimization.}
After initial prediction, each candidate undergoes two refinement
steps (Figure~\ref{fig:data_pipeline}, green):
(1)~\emph{translation optimization}, which aligns the predicted depth
to the estimated depth map using percentile-based scaling or
anchor-based optimization; and
(2)~\emph{rotation optimization}, which corrects orientation using
PCA-based gravity alignment and 2D projection constraints.
The candidates from all five models are then merged into a unified
10D format (center, dimensions, quaternion), yielding up to five
candidates per 2D annotation.

\subsection{Rule-based filtering}
\label{subsec:quality_control}

All candidates and annotations undergo multi-stage filtering.
Failed annotations are never deleted but flagged as \texttt{ignore3D=1},
preserving the full 2D annotation set for recall evaluation.

\paragraph{Rule-based filtering.}
Before selection, candidates are filtered by three geometric criteria:
edge contact ratio $\geq$\,3\% (box at image boundary),
occlusion ratio $>$\,15\% (for the RANSAC-PCA method), or
3D-to-2D projection size ratio outside $[0.5, 1.5]$.
Candidates failing any criterion are discarded before entering the
annotation or VLM selection stage.

\paragraph{Depicted object filter.}
A VLM-based classifier based on Qwen3.5-9B~\cite{bai2025qwen3vltechnicalreport} identifies and discards annotations of \emph{depicted} objects such as pictures, posters, reflections, or screen displays that portray objects rather than real 3D instances.

\paragraph{Composite image filter.}
Images composed of multiple sub-images are detected using
Qwen3.5-9B and removed from the
dataset, since they often result in inaccurate depth maps that could
be confusing for the model.

\paragraph{LLM-estimated size and geometry filter.}
We use GPT-4.1-mini~\cite{openai2024gpt4technicalreport} to estimate expected physical
dimensions for each object category (shortest/middle/longest axis
ranges, depth-to-width ratio bounds, and a fixed/variable size
classification).  Annotations are then filtered in three passes:
\begin{itemize}[nosep]
\item \textbf{Absolute size}: Each axis must fall within the
  LLM-estimated range.  Fixed-size categories (\eg, person, car)
  use a tight tolerance of 1.5$\times$, while variable-size
  categories (\eg, toy, sculpture) use 3.0$\times$ to accommodate
  natural size variation; both are relaxed to 2.5$\times$/5.0$\times$
  for fine-grained datasets.
  Flat and elongated objects skip certain axes.
\item \textbf{Depth-to-width ratio}: The Z/X extent ratio must not
  exceed a per-category maximum, catching depth estimation
  stretching artifacts.
\item \textbf{Axis proportion}: For non-flat, non-elongated objects,
  the shortest-to-middle axis ratio must be plausible.
\end{itemize}

\paragraph{Small object upgrade.}
Objects initially filtered as ``small'' (2D area $<$\,0.5\% of image)
are re-evaluated using the same VLM criteria as synthetic selection.
Qualifying small objects are upgraded to valid annotations, recovering
additional long-tail supervision.

\subsection{Candidate selection}
\label{subsec:annotation}

We obtain final 3D annotations from generated candidates through two complementary paths: human annotation for a carefully sampled subset and VLM-based automatic
selection for the remainder.

\paragraph{Human selection.}
For a balanced subset of images, crowdsourced annotators on
Prolific~\cite{prolific} evaluates up to five candidates per object.
Each candidate is visualized from four viewpoints: a perspective
overlay on the original image and three orthographic point cloud
views.  Annotators \emph{select the best candidate} and \emph{rate
its quality} as \texttt{good\_fit}, \texttt{acceptable}, or
\texttt{unacceptable}.  Each batch contains 50~regular tasks plus
5~gold (quality-control) tasks with known-bad annotations; batches
where annotators fail to identify $\geq$\,2/5 gold tasks are
discarded and reassigned to new qualified annotators until the
target data quality is reached.  Overall pass rates range from
84\% to 98\% across dataset splits.

\paragraph{VLM selection.}
For images without human annotation, we automatically select the best candidate  a Molmo2~\cite{clark2026molmo2openweightsdata} checkpoint fine-tuned for this task with synthetically generated positive and negative candidate pairs from Omni3D. Each candidate is scored on six perceptual criteria: \emph{category correctness}, \emph{scale accuracy}, \emph{translation accuracy}, \emph{shape fidelity}, \emph{rotation correctness}, and \emph{vertical tilt alignment}. Scores range from 0 to 2 for all criteria except category correctness, which ranges from 0 to 1, giving a maximum total score of 11. Given a cropped image overlaid with the projected 3D box wireframe, the fine-tuned Molmo2 predicts structured scores for each criterion. We keep the highest-scoring candidate when its total score is greater than 10.

\subsection{Statistics}
\label{subsec:data_stats}

\begin{table*}[t]
\centering
\caption{
  \textbf{\data statistics.} 
  Human annotations are rated by crowd-source workers, while synthetic annotations are auto-selected by VLM scoring.
  Combined, the dataset spans 13.5K categories---a 138$\times$ increase over Omni3D's 98 categories.
}
\label{tab:data_stats}
\vspace{-0.5em}
\tablestyle{4pt}{1.15}
\resizebox{\linewidth}{!}{%
\begin{tabular}{l l r r r c c c}
Split & Source & Images & Ann.
  & Categories & Type & Scene & Max depth \\
\midrule
\multicolumn{8}{l}{\emph{Existing datasets}} \\
Omni3D~\cite{brazil2023omni3dlargebenchmarkmodel}
  & KITTI, nuSc., SUNRGBD, etc.
  & 234K & 3M+ & 98 & Human & Driving, Furniture & 67\,m \\
COCO-3D~\cite{yao2026labelany3dlabelobject3d}
  & COCO
  & 18K & 92K & 80 & Synthetic & In-the-wild & 35\,m \\
CA-1M~\cite{lazarow2024cubifyanythingscalingindoor}
  & ARKitScenes
  & 3,500 (videos) & 400K & Class-agnostic & Human & Indoor & 5\,m \\
\midrule
\multicolumn{8}{l}{\emph{\data}} \\
Train (Human)
  & COCO, LVIS, Obj365, V3Det
  & 102,979 & 229,934 & 12,064 & Human & In-the-wild & \\
Train (Synthetic)
  & COCO, LVIS, Obj365, V3Det
  & 896,004 & 3,483,292 & 11,896 & VLM filter & In-the-wild & \\
Val
  & COCO, LVIS, Obj365
  & 2,470 & 9,256 & 785 & Human & In-the-wild & \\
Test
  & COCO, LVIS, Obj365
  & 2,433 & 5,596 & 633 & Human & In-the-wild & \\
\midrule
{\color{molmocolor}\data (total)}
  & & \textbf{1,003,886}
  & \textbf{3,728,078} & \textbf{13,499} & Human + VLM & In-the-wild & \textbf{81\,m} \\
\end{tabular}%
}
\end{table*}

Table~\ref{tab:data_stats} summarizes the dataset.  The
human-annotated portion covers $\sim$103K images with quality ratings:
across all splits, 35--48\% of annotations are rated \texttt{good\_fit},
33--50\% \texttt{acceptable}, and 24--39\% \texttt{unacceptable}
(the latter flagged as ignored).  The VLM-filtered portion adds
$\sim$896K images with automatically verified annotations.

\paragraph{Val/test sampling strategy.}
For the validation and test sets, we use a three-phase balanced
sampling algorithm:
(1)~greedy set cover for 100\% category coverage,
(2)~multi-objective balanced fill optimizing category rarity, scene
diversity, depth distribution, and source balance, and
(3)~targeted patching to ensure $\geq$\,3 samples per category.

\paragraph{Category coverage.}
Combining all sources, \data spans \textbf{13,499 unique categories}.  Of the 881 val categories, 826 (99.9\%) have
$\geq$\,1 training annotation, and $\sim$820 have $\geq$\,3.



\paragraph{Candidate model distribution.}
Among valid synthetic annotations, SAM-3D contributes $\sim$55\% of
selected boxes, RANSAC-PCA $\sim$28\%, and LabelAny3D $\sim$17\%,
reflecting the complementary strengths of mesh-based reconstruction,
geometric fitting, and single-image 3D reconstruction.

\paragraph{Filtering impact.}
The multi-stage filtering pipeline removes 15--20\% of annotations via
size/geometry checks, with the absolute size filter contributing the
largest share.  The depiction filter catches $\sim$2\% of annotations
across human and synthetic splits.

\begin{figure*}[t]
\centering
\begin{minipage}[t]{0.65\textwidth}
\centering
\vspace{0pt}
\includegraphics[width=\textwidth]{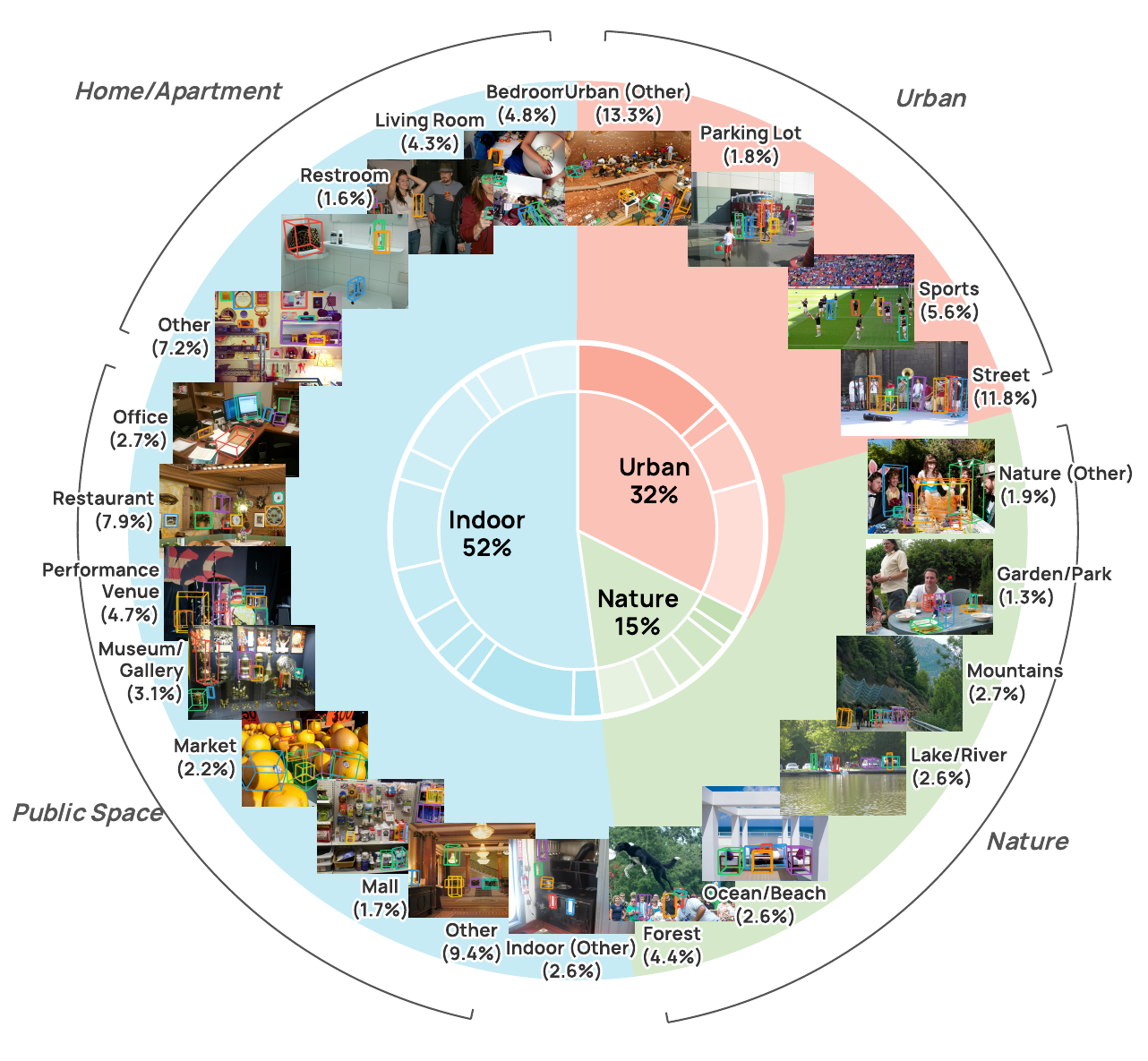}
\caption{\textbf{Scene category distribution of \data.}
Images span three macro-categories: Indoor (52\%), Urban (32\%), and Nature (15\%), with fine-grained sub-categories illustrated by representative examples.}
\label{fig:scene_distribution}
\end{minipage}%
\hfill
\begin{minipage}[t]{0.32\textwidth}
\centering
\vspace{0pt}
\resizebox{\linewidth}{!}{%
\tablestyle{4pt}{1.15}
\begin{tabular}{l c c}
Model & Sel.\ Share & Rej.\ Rate \\
\midrule
SAM-3D     & 40.4\% & 17.3\% \\
RANSAC-PCA & 28.2\% & 12.5\% \\
DetAny3D   & 14.5\% & 42.9\% \\
LabelAny3D & 13.0\% & 21.3\% \\
3D-MOOD    &  3.8\% & 25.7\% \\
\midrule
Overall    &   ---  & 22.0\% \\
\end{tabular}%
}

\vspace{0.6em}

\resizebox{\linewidth}{!}{%
\tablestyle{4pt}{1.15}
\begin{tabular}{c c r}
VLM Score & Rej.\ Rate & \multicolumn{1}{c}{$n$} \\
\midrule
$< 7$ & 71.9\% &   1,992 \\
7     & 67.4\% &  13,670 \\
8     & 45.3\% &  18,665 \\
9     & 36.1\% &  83,882 \\
10    & 16.7\% & 310,329 \\
11    &  9.2\% &  52,684 \\
\midrule
\multicolumn{3}{l}{VLM Top-2 Coverage: \textbf{73.4\%}} \\
\end{tabular}%
}

\vspace{0.3em}
\captionof{table}{
  \textbf{Pipeline validation on the human-annotated train set.}
  \emph{Top:} Candidate model selection share and human rejection rate.
  Rejection rates vary by $>$\,$3\times$ across models.
  \emph{Bottom:} VLM composite score vs.\ human rejection rate.
  Scores correlate perfectly with human judgment.
}
\label{tab:pipeline_validation}
\end{minipage}
\end{figure*}

\subsection{Pipeline validation}
\label{subsec:pipeline_validation}

To validate the annotation pipeline, we analyze the human-annotated
train split (230K accepted annotations) along two axes: how candidate
model quality varies, and how well VLM scoring predicts human judgment.

\paragraph{Candidate model quality.}
Table~\ref{tab:pipeline_validation}~(top) reports the selection share
and rejection rate of each candidate model as judged by human
annotators.  SAM-3D accounts for the largest share of accepted
annotations (40.4\%), followed by RANSAC-PCA (28.2\%), DetAny3D
(14.5\%), LabelAny3D (13.0\%), and 3D-MOOD (3.8\%).  Rejection rates
vary by more than $3\times$ across models: RANSAC-PCA achieves the
lowest rate (12.5\%) while DetAny3D is rejected most frequently
(42.9\%).  This disparity confirms that candidate quality differs
substantially across models and can only be reliably distinguished
through human evaluation.

\paragraph{VLM-human correlation.}
VLM scores exhibit a perfect monotonic correlation with human rejection
rates (Spearman $\rho = -1.0$), as shown in
Table~\ref{tab:pipeline_validation}~(bottom): rejection decreases
steadily from 71.9\% at score $< 7$ to 9.2\% at score~11
(AUC\,=\,0.66, point-biserial $r = 0.30$, $p < 10^{-100}$,
$n = 481$K).  Among the six VLM dimensions, \emph{scale}
(AUC\,=\,0.60) and \emph{shape} (AUC\,=\,0.56) are the strongest
predictors of human acceptance, indicating that size and geometry
fidelity are the primary bottlenecks in candidate quality.
Furthermore, the VLM's top-2 ranked candidates cover \textbf{73.4\%}
of human selections, demonstrating that VLM scoring effectively
narrows the candidate space.

\paragraph{Limits of automatic scoring.}
Despite strong correlation, VLM scoring alone cannot substitute for
human judgment.  Even at score~10---which accounts for 64.5\% of all
candidates (310K)---the human rejection rate remains 16.7\%.  This
gap motivates our two-stage design: VLM scoring as an efficient
pre-filter, followed by human verification for quality-critical
subsets.


\section{Experiments}
\label{sec:experiments}

We first evaluate on \bench, our proposed open-vocabulary
in-the-wild benchmark with 700+ categories
(Section~\ref{subsec:in_the_wild}), then on the standard Omni3D
benchmark (Section~\ref{subsec:main_results}) and
zero-shot transfer to Argoverse~2 and ScanNet
(Section~\ref{subsec:open_set}).
We further test with real depth on Stereo4D
(Section~\ref{subsec:stereo4d}) and present ablation studies
(Section~\ref{subsec:ablation}).
All evaluations use both text-prompt and oracle (box-prompt) modes.

\subsection{Experimental setup}
\label{subsec:setup}

\paragraph{Datasets.}
We evaluate across four benchmarks covering diverse detection settings.
\begin{itemize}[nosep]

\item \textbf{In-the-wild detection on \bench}, which is our proposed in-the-wild benchmark as detailed in Section~\ref{subsec:in_the_wild}.
\bench covers 700+ open-vocabulary categories with human-verified 3D annotations.

\item \textbf{Evaluation on Omni3D}~\cite{brazil2023omni3dlargebenchmarkmodel} unifies six datasets spanning indoor and outdoor scenes:
KITTI~\cite{Geiger2012CVPR} (7.5K outdoor driving images),
nuScenes~\cite{caesar2020nuscenesmultimodaldatasetautonomous} (28K multi-view driving images),
SUNRGBD~\cite{7298655} (10K indoor RGB-D images),
Hypersim~\cite{roberts:2021} (100K+ synthetic indoor images),
ARKitScenes~\cite{baruch2022arkitscenesdiverserealworlddataset} (55K mobile AR indoor images),
and Objectron~\cite{objectron2021} (15K+ object-centric images),
with a unified label space of 98 categories and standardized 3D box annotations.

\item \textbf{Zero-shot detection evaluation.}
Following 3D-MOOD~\cite{yang20253dmoodlifting2d3d}, we evaluate cross-dataset
generalization on Argoverse~2~\cite{wilson2023argoverse2generationdatasets}
(outdoor driving, 26~classes) and ScanNet~\cite{dai2017scannet}
(indoor, 18~classes), both unseen during training.

\item \textbf{Real stereo depth detection on Stereo4D}~\cite{jin2025stereo4d}, which provides 383 in-the-wild images
with real depth maps across 78 categories, used to evaluate
depth generalization zero-shot.

\end{itemize}

\paragraph{Evaluation metrics.}
We evaluate in two modes:
(1)~\textbf{text prompt}, where category names are used as open-vocabulary text queries for end-to-end detection, and
(2)~\textbf{box prompt} (oracle), where ground-truth 2D boxes serve as geometric prompts to isolate 3D regression quality.
For \textbf{Omni3D}, we follow the standard protocol and report
Average Precision (AP$_\text{3D}$) at 3D~IoU thresholds $[0.05\!:\!0.50\!:\!0.05]$
(10 thresholds).
Objects annotated as \emph{ignore} (\eg, invalid 3D annotation,
heavy truncation) are excluded from both the positive ground-truth
set and the false-positive count: a prediction matching an ignored
object is treated as neutral (see Section~\ref{subsec:loss}).
For \textbf{zero-shot transfer} on Argoverse~2 and ScanNet, we
follow the 3D-MOOD protocol~\cite{yang20253dmoodlifting2d3d} and
report the Open Detection Score (ODS), which combines
AP with three error metrics into a unified score:
$\text{ODS} = (3\!\cdot\!\text{AP} + (1\!-\!\text{mATE}) + (1\!-\!\text{mAOE}) + (1\!-\!\text{mASE})) / 6$,
where mATE, mAOE, and mASE denote mean translation, orientation,
and scale errors, respectively.
We use the canonical rotation convention for mAOE, which resolves
the $180^\circ$ ambiguity for symmetric objects.

\paragraph{In-the-wild evaluation protocol.}
From the validation set, we construct \bench, an in-the-wild evaluation
benchmark spanning 700+ open-vocabulary categories from COCO, LVIS,
and Objects365.  Categories are split by annotation frequency into
three groups:
\textbf{rare} ($<$\,5 samples, 464~categories),
\textbf{common} (5--20 samples, 283~categories), and
\textbf{frequent} ($>$\,20 samples, 63~categories).
For \bench and Stereo4D, where open-vocabulary categories span a
wide range of object sizes, we report AP using center-distance
matching~\cite{yang20253dmoodlifting2d3d} (AP$_\text{3D}$):
a prediction is matched to a ground truth if the 3D center distance
is within a fraction of the object radius, with thresholds
$[0.50\!:\!1.00\!:\!0.05]$.
We report AP$_\text{3D}$ separately for each frequency group
(AP$_\text{rare}$, AP$_\text{common}$, AP$_\text{frequent}$ ) and overall.
Since the annotations are not exhaustive---not every object in each
image has a valid 3D bounding box---we follow the federated evaluation
protocol of LVIS~\cite{gupta2019lvis}: for text-prompt evaluation,
a prediction that overlaps with a 2D-annotated object lacking a valid
3D box is treated as neutral rather than a false positive.

\paragraph{Implementation details.}
We train \model in three stages using AdamW~\cite{loshchilov2019decoupledweightdecayregularization}
with a base learning rate of $10^{-4}$ and weight decay of $10^{-4}$,
on 4 nodes (32~GPUs) with per-GPU batch size 4 (effective batch size 128).
\begin{itemize}[nosep]
\item \textbf{Stage~1} trains on Omni3D for 12~epochs.
\item \textbf{Stage~2} fine-tunes on a mixture of Omni3D and \data (human and synthetic), plus several supplementary 3D datasets collectively referred to as ``Others'': CA-1M~\cite{lazarow2024cubifyanythingscalingindoor} (indoor), Waymo~\cite{Sun_2020_CVPR} (driving), 3EED~\cite{li20253eedground3d} (detection and referring expression), and FoundationPose~\cite{wen2024foundationposeunified6dpose} (object-level pose). These are limited to closed-set categories and specific domains but provide complementary depth ranges and scene layouts that strengthen the model's geometric estimation; open-vocabulary diversity is primarily driven by \data. Stage~2 trains for 12~epochs.
\item \textbf{Stage~3} further fine-tunes on Omni3D and \data (human) with
mask-guided point/box training for 3~epochs.
The learning rate follows a multi-step decay schedule; full data
mixing ratios and schedule details are in the appendix.
\end{itemize}
The SAM3 backbone is partially frozen (first 28 transformer blocks fixed).
The LingBot-Depth geometry backend encoder is also partially frozen
(first 21 of 24 blocks fixed, last 3 trainable).
The 3D detection head is trained from scratch.
Input images are resized to $1008\!\times\!1008$ pixels.
Data augmentation includes random resizing (scale $[0.75,\,1.25]$)
and random horizontal flip.
At test time, we apply per-category NMS with an IoU threshold of~0.6.

\subsection{In-the-wild evaluation: Results on WildDet3D-Bench}
\label{subsec:in_the_wild}

We evaluate on \bench (Section~\ref{subsec:in_the_wild}), which
covers 700+ open-vocabulary categories with human-verified 3D
annotations.

\begin{table*}[t]
\centering
\caption{
  \textbf{\bench evaluation}.
We observe that (1) \model outperforms baseline models when trained on the same data (\ie, Omni3D), (2) our newly introduced \data further improves performance by a significant margin (6.8$\rightarrow$22.6, 8.4$\rightarrow$24.8), and (3) incorporating depth input at test time nearly doubles performance (22.6$\rightarrow$41.6, 24.8$\rightarrow$47.2).
}
\label{tab:in_the_wild}
\vspace{-0.5em}
\tablestyle{4pt}{1.15}
\begin{tabular}{l l  | c c c | c}
Method & Training data
  & AP$_\text{rare}$ & AP$_\text{common}$ & AP$_\text{frequent}$
  & AP$_\text{3D}$ \\
\midrule
\multicolumn{2}{l|}{\emph{Text Prompt}} & \multicolumn{3}{l|}{} & \\
3D-MOOD~\cite{yang20253dmoodlifting2d3d}
  & Omni3D
  & 2.4 & 2.1 & 2.6
  & 2.3 \\
{\color{molmocolor}\model}
  & Omni3D
  & 9.0 & 6.5 & 5.2
  & 6.8 \\
{\color{molmocolor}\model w/ depth}
  & Omni3D
  & 23.0 & 21.5 & 16.1
  & 20.7 \\
{\color{molmocolor}\model}
  & Omni3D, Others, \data
  & \underline{28.3} & \underline{21.6} & \underline{18.7}
  & \underline{22.6} \\
{\color{molmocolor}\model w/ depth}
  & Omni3D, Others, \data
  & \textbf{47.4} & \textbf{40.7} & \textbf{37.2}
  & \textbf{41.6} \\
\midrule
\multicolumn{2}{l|}{\emph{Box Prompt }} & \multicolumn{3}{l|}{} & \\
OVMono3D-LIFT~\cite{yao2025openvocabularymonocular3d}
  & Omni3D
  & 7.4 & 8.8 & 5.1
  & 7.7 \\
DetAny3D~\cite{zhang2025detect3dwild}
  & Omni3D, Others
  & 9.9 & 7.4 & 6.3
  & 7.8 \\
{\color{molmocolor}\model}
  & Omni3D
  & 12.0 & 7.9 & 5.3
  & 8.4 \\
{\color{molmocolor}\model w/ depth}
  & Omni3D
  & 26.4 & \underline{24.4} & 19.6
  & 23.9 \\
{\color{molmocolor}\model}
  & Omni3D, Others, \data
  & \underline{30.0} & 24.2 & \underline{20.3}
  & \underline{24.8} \\
{\color{molmocolor}\model w/ depth}
  & Omni3D, Others, \data
  & \textbf{53.7} & \textbf{46.1} & \textbf{42.5}
  & \textbf{47.2} \\
\end{tabular}
\end{table*}

\paragraph{Results.}
Table~\ref{tab:in_the_wild} presents comprehensive in-the-wild results
across prompt modes and depth settings.
When trained on Omni3D only with text prompts, our method already
achieves 6.8~AP, outperforming 3D-MOOD (2.3~AP) by
$\mathbf{3.0\times}$.
With additional training data (+ Others + \data), our text-prompt result reaches \textbf{22.6~AP}---a $\mathbf{9.8\times}$ improvement over 3D-MOOD.

\paragraph{Effect of GT depth.}
Providing ground-truth depth at test time dramatically improves
performance.  For the Omni3D-only model, text-prompt AP jumps from
6.8 to 20.7 ($\mathbf{+13.9}$); for the full model, it increases
from 22.6 to \textbf{41.6} ($\mathbf{+19.0}$), demonstrating that
our architecture effectively leverages depth signals when available.

\paragraph{Text \vs box prompt.}
Box prompts consistently outperform text prompts when no depth
is provided (8.4 \vs 6.8 for Omni3D; 24.8 \vs 22.6 for +Others+\data),
confirming that 2D detection is a bottleneck.
Interestingly, with GT depth the text-prompt setting (41.6~AP)
is competitive with oracle (47.2~AP) for the full model, and both
dramatically outperform all baselines.

\paragraph{Frequency splits.}
The improvements are consistent across all frequency groups, with the
largest gains on rare categories (AP$_\text{rare}$\,=\,47.4 \vs 2.4 for
3D-MOOD), demonstrating strong generalization to novel categories.

\subsection{Results on Omni3D}
\label{subsec:main_results}

We compare our method against several baselines:
Cube R-CNN~\cite{brazil2023omni3dlargebenchmarkmodel}, a strong monocular 3D detector;
Uni-MODE~\cite{Li_2024_CVPR}, a unified monocular 3D detection model;
3D-MOOD~\cite{yang20253dmoodlifting2d3d} with Swin-T and Swin-B backbones; and
DetAny3D~\cite{zhang2025detect3dwild}, a recent 3D detection foundation model.

\begin{table*}[t]
\centering
\caption{
\textbf{Omni3D evaluation.}  
Our model outperforms previous state-of-the-art methods in both text and box prompt settings. Incorporating depth input at test time further improves performance significantly.
}
\label{tab:omni3d}
\vspace{-0.5em}
\tablestyle{4pt}{1.15}
\resizebox{\linewidth}{!}{%
\begin{tabular}{l | c c c c c c | c}
Method & KITTI~\cite{Geiger2012CVPR} & nuScenes~\cite{caesar2020nuscenesmultimodaldatasetautonomous} & SUNRGBD~\cite{7298655} & Hypersim~\cite{roberts:2021} & ARKitScenes~\cite{baruch2022arkitscenesdiverserealworlddataset} & Objectron~\cite{objectron2021}
  & AP$_\text{3D}$ \\
\midrule
\multicolumn{1}{l|}{\emph{Text Prompt}} & \multicolumn{6}{l|}{} & \\
Cube R-CNN~\cite{brazil2023omni3dlargebenchmarkmodel}
  & 32.6 & 30.1 & 15.3 & 7.5 & 41.7 & 50.8
  & 23.3 \\
Uni-MODE*~\cite{Li_2024_CVPR}
  & 29.2 & \textbf{36.0} & 23.0 & 8.1 & 48.0 & 66.1
  & 28.2 \\
3D-MOOD Swin-T~\cite{yang20253dmoodlifting2d3d}
  & 32.8 & 31.5 & 21.9 & 10.5 & 51.0 & 64.3
  & 28.4 \\
3D-MOOD Swin-B~\cite{yang20253dmoodlifting2d3d}
  & 31.4 & \underline{35.8} & 23.8 & 9.1 & 53.9 & \underline{67.9}
  & 30.0 \\
{\color{molmocolor}\model}
  & \textbf{37.0} & 31.7 & \underline{38.9} & \underline{16.5} & \underline{64.6} & 60.5
  & \underline{34.2} \\
{\color{molmocolor}\model w/ depth}
  & \underline{36.1} & 32.0 & \textbf{51.1} & \textbf{26.6}
  & \textbf{73.3} & \textbf{68.3}
  & \textbf{41.6} \\
\midrule
\multicolumn{1}{l|}{\emph{Box Prompt }} & \multicolumn{6}{l|}{} & \\
OVMono3D-LIFT~\cite{yao2025openvocabularymonocular3d}
  & 31.4 & 32.5 & 23.2 & 11.9 & 54.2 & \underline{63.5}
  & 29.6 \\
DetAny3D~\cite{zhang2025detect3dwild}
  & 38.7 & \textbf{37.6} & \underline{46.1} & 16.0 & 50.6 & 56.8
  & 34.4 \\
{\color{molmocolor}\model}
  & \textbf{44.3} & 35.3 & 43.1 & \underline{17.3} & \underline{66.6} & 60.8
  & \underline{36.4} \\
{\color{molmocolor}\model w/ depth}
  & \underline{42.8} & \underline{35.9} & \textbf{58.7} & \textbf{30.4}
  & \textbf{76.6} & \textbf{68.5}
  & \textbf{45.8} \\
\end{tabular}%
}
\end{table*}

\paragraph{Results.}
Table~\ref{tab:omni3d} reports results on Omni3D in both text prompt
and oracle (box prompt) settings.
With text prompts, our method achieves 34.2~AP, surpassing 3D-MOOD
(28.4~AP) by \textbf{+5.8~AP} with $10\times$ fewer training epochs
(12 \vs 120).  In the oracle setting, our method reaches 36.4~AP,
outperforming DetAny3D (34.4~AP) by \textbf{+2.0~AP} despite
training for only 12~epochs \vs 80.  The gains are especially
pronounced on indoor datasets: ARKitScenes and Objectron, demonstrating
stronger geometry estimation in cluttered environments.

\paragraph{Effect of sparse depth.}
When sparse depth is provided at test time, performance further improves
across both settings: oracle with depth reaches \textbf{45.8~AP}
(\textbf{+11.4} over DetAny3D), with dramatic gains on indoor datasets
equipped with depth sensors (SUNRGBD, Hypersim, ARKitScenes).

\paragraph{Training efficiency.}
A key advantage of our approach is training efficiency: we achieve
superior results with 12~epochs compared to 80--120~epochs for
baselines.  This is enabled by the strong pre-trained representations
from SAM3 and LingBot-Depth, which provide a high-quality
initialization for both detection and depth estimation.

\subsection{Zero-shot evaluation}
\label{subsec:open_set}

To evaluate cross-dataset generalization, we train on Omni3D and test zero-shot on Argoverse~2~\cite{wilson2023argoverse2generationdatasets} (outdoor driving, 26~classes) and ScanNet~\cite{dai2017scannet} (indoor, 18~classes including novel categories unseen in Omni3D).

\begin{table*}[t]
\centering
\caption{
  \textbf{Zero-shot evaluation.} 
  ODS is the Open Detection Score~\cite{yang20253dmoodlifting2d3d}; higher is better.
  mATE, mASE, mAOE denote mean translation, scale, and orientation errors (lower is better).
}
\label{tab:open_set}
\vspace{-0.5em}
\tablestyle{3pt}{1.15}
\small
\begin{tabular}{l | c c c c c | c c c c c}
 & \multicolumn{5}{c|}{Argoverse 2~\cite{wilson2023argoverse2generationdatasets}}
  & \multicolumn{5}{c}{ScanNet~\cite{dai2017scannet}} \\
Method
  & AP$\uparrow$ & mATE$\downarrow$ & mASE$\downarrow$ & mAOE$\downarrow$ & ODS$\uparrow$
  & AP$\uparrow$ & mATE$\downarrow$ & mASE$\downarrow$ & mAOE$\downarrow$ & ODS$\uparrow$ \\
\midrule
Cube R-CNN~\cite{brazil2023omni3dlargebenchmarkmodel}
  & 8.6 & 0.903 & 0.867 & 0.953 & 8.9
  & 20.0 & 0.733 & 0.774 & 0.921 & 19.5 \\
3D-MOOD Swin-T~\cite{yang20253dmoodlifting2d3d}
  & 14.8 & 0.782 & 0.697 & 0.612 & 22.5
  & 27.3 & 0.630 & 0.726 & 0.650 & 30.2 \\
3D-MOOD Swin-B~\cite{yang20253dmoodlifting2d3d}
  & 14.7 & 0.755 & 0.680 & 0.580 & 23.8
  & 28.8 & 0.612 & \textbf{0.706} & 0.655 & 31.5 \\
{\color{molmocolor}\model}
  & \underline{43.4} & \underline{0.714} & \underline{0.645} & \underline{0.526} & \underline{40.3}
  & \underline{56.5} & \underline{0.601} & 0.720 & \underline{0.437} & \underline{48.9} \\
{\color{molmocolor}\model w/ depth}
  & \textbf{43.4} & \textbf{0.701} & \textbf{0.645} & \textbf{0.526} & \textbf{40.4}
  & \textbf{57.6} & \textbf{0.589} & \underline{0.707} & \textbf{0.422} & \textbf{50.2} \\
\end{tabular}%
\end{table*}

\paragraph{Results.}
Table~\ref{tab:open_set} shows the zero-shot results.
Our model achieves \textbf{40.3~ODS} on AV2 and
\textbf{48.9~ODS} on ScanNet, outperforming 3D-MOOD Swin-B
by $\mathbf{+16.5}$ and $\mathbf{+17.4}$~ODS respectively.
The detection AP is substantially higher than all baselines:
43.4 \vs 14.8 on AV2 (+28.6) and 56.5 \vs 28.8 on ScanNet (+27.7),
demonstrating strong cross-dataset generalization.
Our model also achieves the best orientation estimation (mAOE):
0.526 on AV2 and 0.437 on ScanNet, significantly better than
3D-MOOD Swin-B (0.580 and 0.655).
On AV2, our model also achieves the best translation accuracy
(mATE\,=\,0.714 \vs 0.755 for Swin-B), showing that the large AP
gain does not come at the cost of localization precision.

\paragraph{Effect of GT depth on cross-dataset transfer.}
Providing ground-truth depth yields a clear improvement on ScanNet
(48.9$\to$\textbf{50.2}~ODS, +1.3), where indoor scenes benefit from
accurate metric depth for resolving scale ambiguity.
On AV2 the gain is marginal (40.3$\to$40.4), suggesting that the
model's monocular depth estimation is already well-calibrated for
outdoor driving scenes at the scale and depth ranges present in
Argoverse~2.

\subsection{In-the-wild evaluation with real depth}
\label{subsec:stereo4d}

To further validate generalization with real depth, we evaluate on Stereo4D~\cite{jin2025stereo4d}, a video dataset with real stereo
depth maps (383 images, 78~categories after filtering), with 2D annotations collected using the SVG2 pipeline~\cite{gao2026syntheticvisualgenome2}. Categories are split into rare ($<$5), common (5--10), and frequent ($\geq$10) groups.
AP is computed using center-distance matching.

\begin{table*}[t]
\centering
\caption{
  \textbf{Stereo4D evaluation.}
  AP is computed using center-distance matching.
  All models are evaluated in a zero-shot manner, \ie, not trained on Stereo4D.
}
\label{tab:stereo4d}
\vspace{-0.5em}
\tablestyle{4pt}{1.15}
\small
\begin{tabular}{l | c c c | c}
Method
  & AP$_\text{rare}$ & AP$_\text{common}$ & AP$_\text{frequent}$
  & AP$_\text{3D}$ \\
\midrule
\multicolumn{1}{l|}{\emph{Box Prompt }} & \multicolumn{3}{l|}{} & \\
OVMono3D-LIFT~\cite{yao2025openvocabularymonocular3d}
  & \underline{12.3} & 7.1 & \underline{11.4}
  & \underline{9.9} \\
DetAny3D~\cite{zhang2025detect3dwild}
  & 8.3 & \underline{8.2} & 4.9
  & 7.1 \\
{\color{molmocolor}\model}
  & 8.1 & 6.3 & 8.5
  & 7.5 \\
{\color{molmocolor}\model w/ depth}
  & \textbf{26.2} & \textbf{31.1} & \textbf{24.6}
  & \textbf{27.7} \\
\end{tabular}
\end{table*}

\paragraph{Results.}
Table~\ref{tab:stereo4d} shows zero-shot results on Stereo4D.
Without depth, our monocular model (7.5~AP) is competitive with
DetAny3D (7.1~AP), while OVMono3D-LIFT achieves the highest monocular
AP (9.9) due to stronger monocular depth estimation on this
low-resolution stereo domain.
When real depth is provided, performance dramatically improves
to \textbf{27.7~AP}, a $2.8\times$ improvement over OVMono3D-LIFT
(9.9~AP), demonstrating that our architecture effectively leverages
real depth signals for accurate 3D localization.

\subsection{Ablation}
\label{subsec:ablation}

We conduct ablation studies to analyze the contribution of individual
components.  All ablations use the oracle (box prompt)
evaluation setting on Omni3D, training on Omni3D only.

\begin{table*}[t]
\centering
\caption{
  \textbf{Detection head architecture.}
  Joint 2D+3D prediction is critical; the 3D confidence head
  provides complementary geometry-aware scoring.
  Evaluated in the oracle setting on Omni3D.
}
\label{tab:ablation_arch}
\vspace{-0.5em}
\tablestyle{4pt}{1.15}
\resizebox{\linewidth}{!}{%
\begin{tabular}{l | c c c c c c | c}
Configuration & KITTI~\cite{Geiger2012CVPR} & nuScenes~\cite{caesar2020nuscenesmultimodaldatasetautonomous} & SUNRGBD~\cite{7298655} & Hypersim~\cite{roberts:2021} & ARKitScenes~\cite{baruch2022arkitscenesdiverserealworlddataset} & Objectron~\cite{objectron2021} & AP$_\text{3D}$ \\
\midrule
Full model
  & 27.9 & 28.2 & 33.9 & 13.2 & 59.4 & 56.8
  & 30.2 \\
\quad w/o 3D confidence head
  & 28.0 & 27.9 & 32.1 & 13.0 & 58.2 & 56.9
  & 29.4 {\color{gray}\scriptsize($-$0.8)} \\
\quad w/o 2D head (3D only)
  & 18.3 & 15.6 & 5.1 & 9.7 & 28.5 & 10.9
  & 11.1 {\color{gray}\scriptsize($-$19.1)} \\
\end{tabular}}
\end{table*}

\begin{table*}[b]
\centering
\caption{
  \textbf{Training objectives.}
  One-to-many matching and geometry loss are the most impactful
  training signals; deep supervision and ignore-aware suppression
  provide smaller but consistent gains.
  Evaluated in the oracle setting on Omni3D.
}
\label{tab:ablation_training}
\vspace{-0.5em}
\tablestyle{4pt}{1.15}
\resizebox{\linewidth}{!}{%
\begin{tabular}{l | c c c c c c | c}
Configuration & KITTI~\cite{Geiger2012CVPR} & nuScenes~\cite{caesar2020nuscenesmultimodaldatasetautonomous} & SUNRGBD~\cite{7298655} & Hypersim~\cite{roberts:2021} & ARKitScenes~\cite{baruch2022arkitscenesdiverserealworlddataset} & Objectron~\cite{objectron2021} & AP$_\text{3D}$ \\
\midrule
Full model
  & 27.9 & 28.2 & 33.9 & 13.2 & 59.4 & 56.8
  & 30.2 \\
\quad w/o O2M matching
  & 23.2 & 23.9 & 30.8 & 12.2 & 56.8 & 53.5
  & 27.7 {\color{gray}\scriptsize($-$2.5)} \\
\quad w/o geometry loss
  & 28.3 & 27.7 & 28.6 & 11.1 & 57.0 & 56.4
  & 28.5 {\color{gray}\scriptsize($-$1.7)} \\
\quad w/o deep supervision
  & 28.1 & 28.3 & 32.4 & 12.6 & 58.7 & 56.6
  & 29.9 {\color{gray}\scriptsize($-$0.3)} \\
\quad w/o ignore-aware suppression
  & 28.2 & 29.4 & 33.2 & 13.0 & 59.2 & 56.4
  & 30.0 {\color{gray}\scriptsize($-$0.2)} \\
\end{tabular}}
\end{table*}

\paragraph{Joint 2D+3D detection.}
The most critical architectural choice is joint 2D and 3D prediction
through a shared detection head.
Removing the 2D head and predicting 3D boxes directly causes AP to
collapse from 30.2 to 11.1 ($-$19.1), with indoor datasets hit
hardest (SUNRGBD: 33.9$\to$5.1, Objectron: 56.8$\to$10.9).
This confirms that 2D detection provides essential spatial priors---
accurate object localization in the image plane---that anchor the
subsequent 3D regression.
Without these priors, the model struggles to jointly localize and
estimate 3D geometry from scratch.

\paragraph{3D confidence head.}
The 3D confidence head (Eq.~\ref{eq:3d_conf}) re-ranks detections
using a geometry-aware score that complements 2D objectness.
Removing it drops AP by 0.8 (30.2$\to$29.4), since 2D objectness
alone cannot distinguish well-localized 3D predictions from
spatially inaccurate ones.

\paragraph{One-to-many matching.}
Among training objectives, one-to-many (O2M) auxiliary matching
contributes the largest gain ($-$2.5~AP without it).
The drop is most pronounced on driving datasets
(KITTI: 27.9$\to$23.2, nuScenes: 28.2$\to$23.9), where dense,
similarly-sized objects benefit most from the richer supervision
signal that O2M provides during training.

\paragraph{Geometry loss.}
Explicit geometric supervision through the depth and camera-ray
losses contributes $-$1.7~AP.  The effect concentrates on indoor
scenes (SUNRGBD: 33.9$\to$28.6, Hypersim: 13.2$\to$11.1), where
accurate metric depth estimation is essential for correct 3D box
placement.

\paragraph{Deep supervision and ignore-aware suppression.}
Deep supervision on intermediate decoder layers provides a modest
$-$0.3~AP improvement.
Ignore-aware suppression---which prevents the model from being
penalized for detecting objects marked as ignore during
evaluation---contributes $-$0.2~AP on Omni3D.
The small magnitude is expected since ignore annotations are sparse
in this benchmark; we expect a larger effect on \bench where
partial 3D annotations are common.

\paragraph{Sparse depth at test time.}
When sparse depth measurements from depth sensors are available at
inference, they can be incorporated through the geometry backend.
As shown in Table~\ref{tab:omni3d}, this provides
substantial gains, particularly on indoor datasets equipped with
RGB-D sensors.  The improvement of +9.4~AP (oracle: 36.4$\to$45.8) and +7.4~AP
(text: 34.2$\to$41.6) demonstrates that our architecture gracefully
accommodates additional depth signals without architectural changes.

\begin{figure*}[!b]
  \centering
  \includegraphics[width=\textwidth]{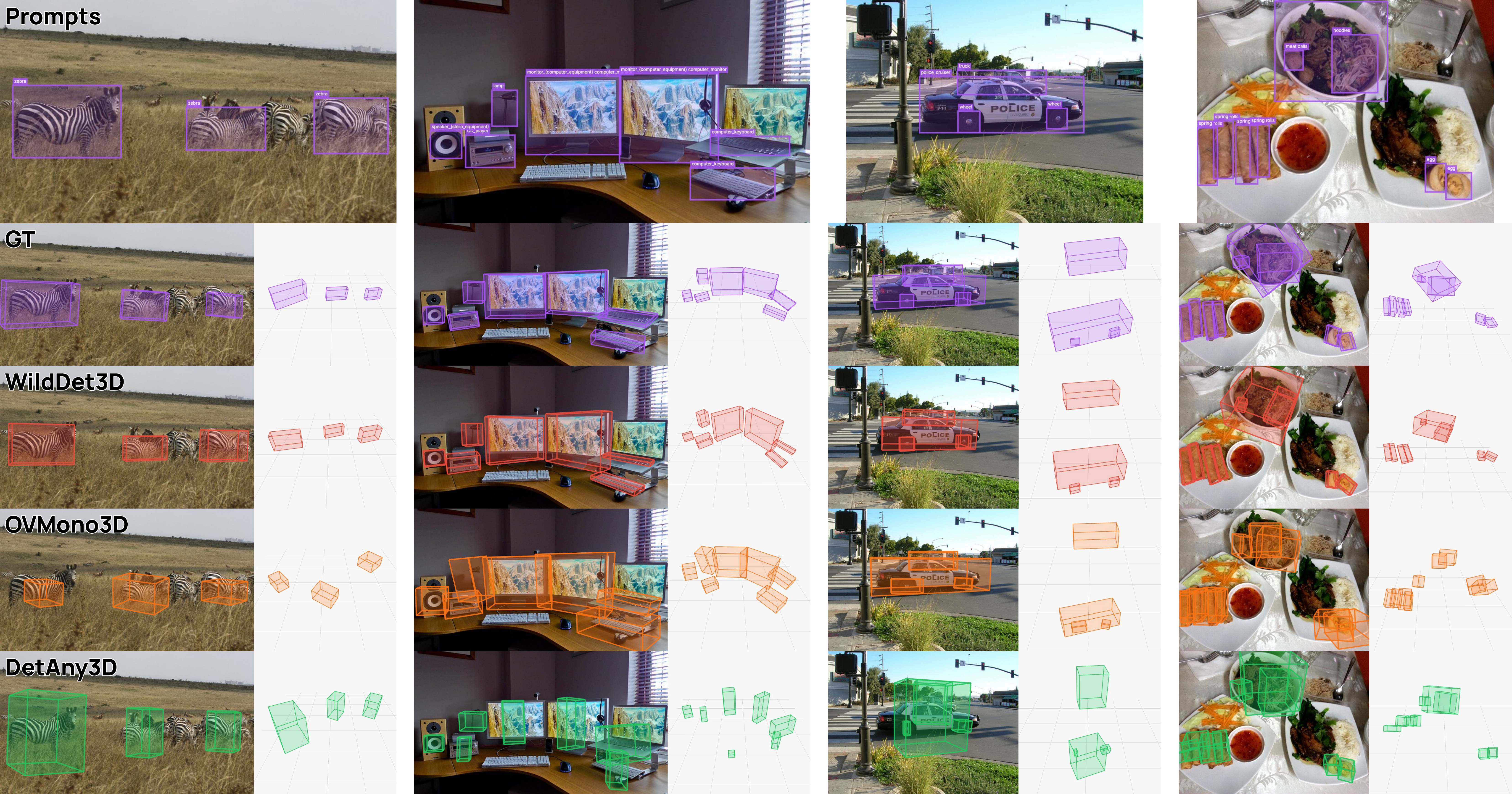}
  \caption{\textbf{Qualitative comparison on in-the-wild images (box prompts).}
  Each block shows the same scene detected by three models, all prompted using 2D bounding boxes. From top to bottom: 2D box prompt visualizations (only box prompts are used, the text labels are for reference), ground truth 3D boxes, \model predictions, OVMono3D predictions, and DetAny3D predictions, with 2D overlays and corresponding 3D bounding boxes.
  \model produces more accurate 3D localization and tighter boxes across diverse scenarios, including animals, vehicles, indoor electronics, and small food items.}
  \label{fig:model_comparison_box}
\end{figure*}

\subsection{Qualitative results}
\label{subsec:qualitative}

Figure~\ref{fig:model_comparison_box} compares \model against OVMono3D~\cite{yao2025openvocabularymonocular3d} and DetAny3D~\cite{zhang2025detect3dwild} on four in-the-wild scenes using box prompts.
On the outdoor animal scene (first scene), \model correctly localizes multiple zebras with tight 3D boxes, while DetAny3D and OVMono3D both detect unrealistic bounding box shapes.
On the cluttered indoor desk (second scene), \model detects fine-grained objects such as monitors and keyboards with correct scale and orientation, whereas other methods struggle with overlapping objects and inaccurate bounding boxes.
For the street scene (third scene), \model accurately captures objects at varying depths; competing methods either hallucinate large boxes (DetAny3D) or fail to estimate correct orientation (OVMono3D).
On the food scene (fourth scene), \model produces well-fitting 3D boxes for individual dishes at close range, with accurate inter-object bounding box relationships (the noodles and the meatballs bounding box fit entirely inside the bowl bounding box), while competing models predict unrealistically placed objects or incorrect dimensions.

\begin{figure*}[t]
  \centering
  \includegraphics[width=\textwidth]{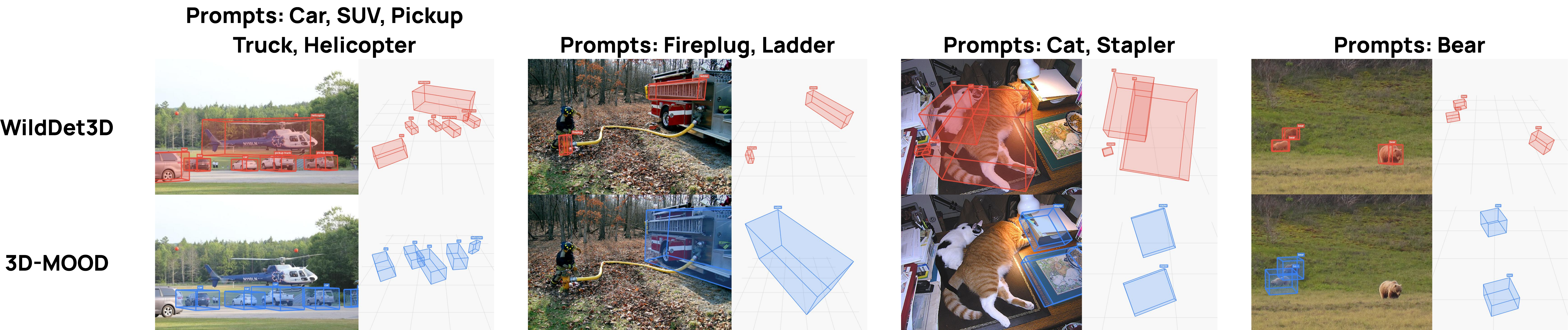}
  \caption{\textbf{Qualitative comparison on in-the-wild images (text prompts).}
  Each block shows the same scene detected by WildDet3D (top) and 3D-MOOD (bottom), prompted with text categories only.}
  \label{fig:model_comparison_text}
\end{figure*}

Figure~\ref{fig:model_comparison_text} compares \model against 3D-MOOD~\cite{yang20253dmoodlifting2d3d} on four in-the-wild scenes using text prompts.
\model consistently detects more object categories and more realistic object placement orientation, relative position, and shape.

These results highlight \model's advantages in multi-object scenes with diverse scales, cluttered layouts, and open-vocabulary categories. For more qualitative results see Appendix~\ref{appendix:qualitative}.

\section{Applications}
\label{sec:application}

Beyond benchmark evaluation, we demonstrate \model across a range of real-world deployment scenarios, spanning on-device mobile inference, AR headsets, vision-language model integration, and robotic manipulation.

\begin{figure*}[!b]
  \centering
  \begin{subfigure}[b]{\textwidth}
    \centering
    \includegraphics[width=\textwidth]{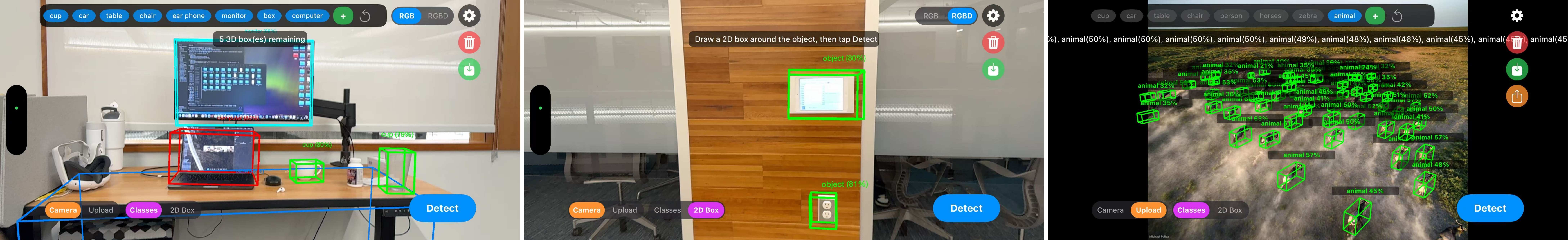}
    \caption{Mobile APP (iPhone) : text prompt detection in an office (left), 2D box prompt for geometric detection (middle), and open-vocabulary animal detection outdoors (right).}
    \label{fig:iphone_demo}
  \end{subfigure}
  \vspace{0.3em}
  \begin{subfigure}[b]{\textwidth}
    \centering
    \includegraphics[width=\textwidth]{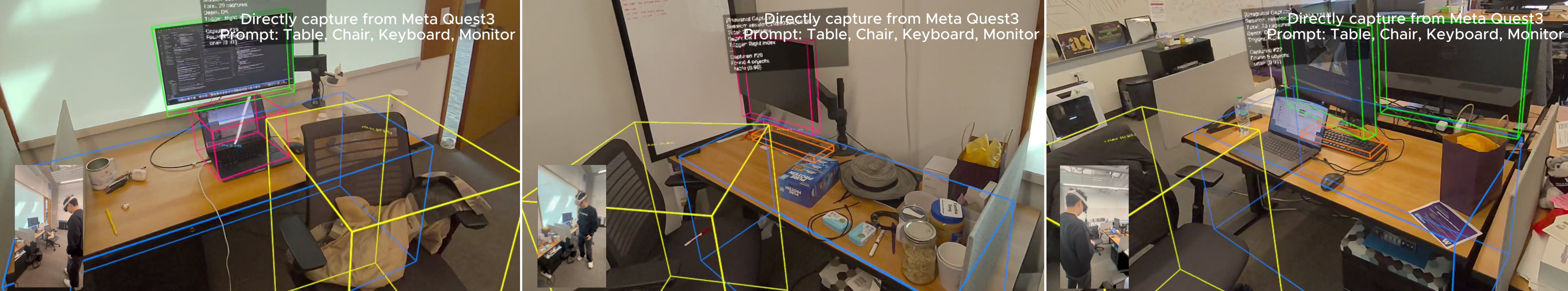}
    \caption{Augmented Reality (Meta Quest 3): passthrough AR with 3D bounding boxes rendered in real time across three different desk scenes.}
    \label{fig:quest_demo}
  \end{subfigure}
  \vspace{0.3em}
  \begin{subfigure}[b]{\textwidth}
    \centering
    \includegraphics[width=\textwidth]{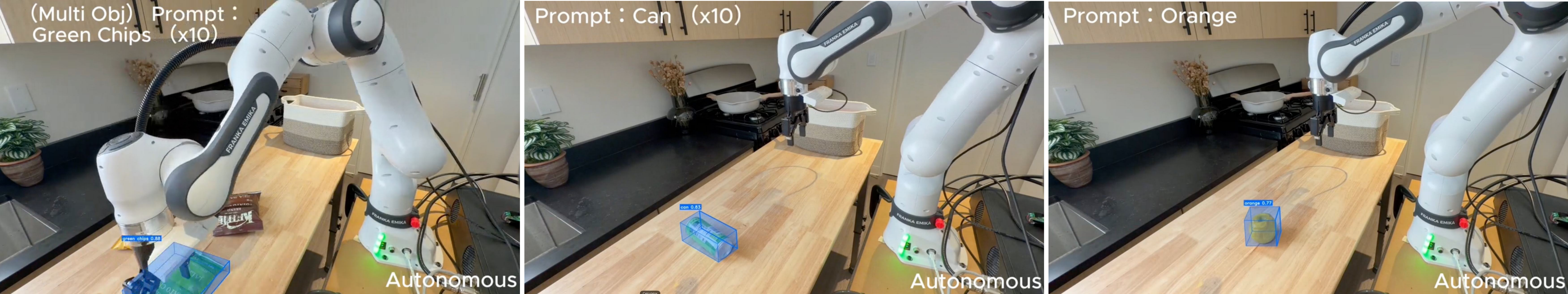}
    \caption{Robotics (manipulation): Franka Emika Panda autonomously grasping objects specified by open-vocabulary text prompts (``Green Chips'', ``Can'', ``Orange'').}
    \label{fig:robotics_demo}
  \end{subfigure}
  \caption{\textbf{Real-world deployment demos.} Each row shows three frames from a different deployment platform, demonstrating \model across diverse interaction modes and environments.
  }
  \label{fig:app_demos}
\end{figure*}

\paragraph{\model in your pocket.}
We deploy \model on iPhone via a client-server architecture, where the iPhone captures RGB frames and LiDAR depth via ARKit and streams them to a cloud-hosted inference endpoint (Figure~\ref{fig:iphone_demo}).
The APP supports multiple interaction modes: open-vocabulary text queries, 2D bounding box prompts for geometric detection, and real-time camera-based inference.
Detected 3D boxes are rendered as AR overlays anchored to the physical scene using ARKit's world tracking, enabling interactive 3D perception on consumer hardware.
The app is publicly available on the App Store.

\paragraph{\model for Augmented Reality (AR).}
We integrate \model with Meta Quest 3 to enable 3D object detection in augmented reality (Figure~\ref{fig:quest_demo}).
The Unity client captures passthrough camera frames with calibrated intrinsics and 6-DoF pose from the Quest's tracking system, sends them to the \model API, and renders detected 3D bounding boxes as overlays in the passthrough view.
This enables spatial understanding for AR applications, where users can query objects in their environment by category and see metric 3D boxes anchored in physical space.


\paragraph{\model for Robotics.}
We apply \model to robotic manipulation with a Franka Emika Panda arm (Figure~\ref{fig:robotics_demo}), where accurate 3D localization is essential for grasping and interaction planning.
A third-view camera captures the scene, and \model produces open-vocabulary 3D bounding boxes that are transformed into the robot's coordinate frame.
The predicted box centers and dimensions are directly consumed for grasp pose generation, which will input to an IK-based interpolation planner for robot to execute, providing a zero-shot alternative to task-specific 3D perception modules that require per-object training or CAD models.

\begin{figure}[!b]
  \centering
  \includegraphics[width=\linewidth]{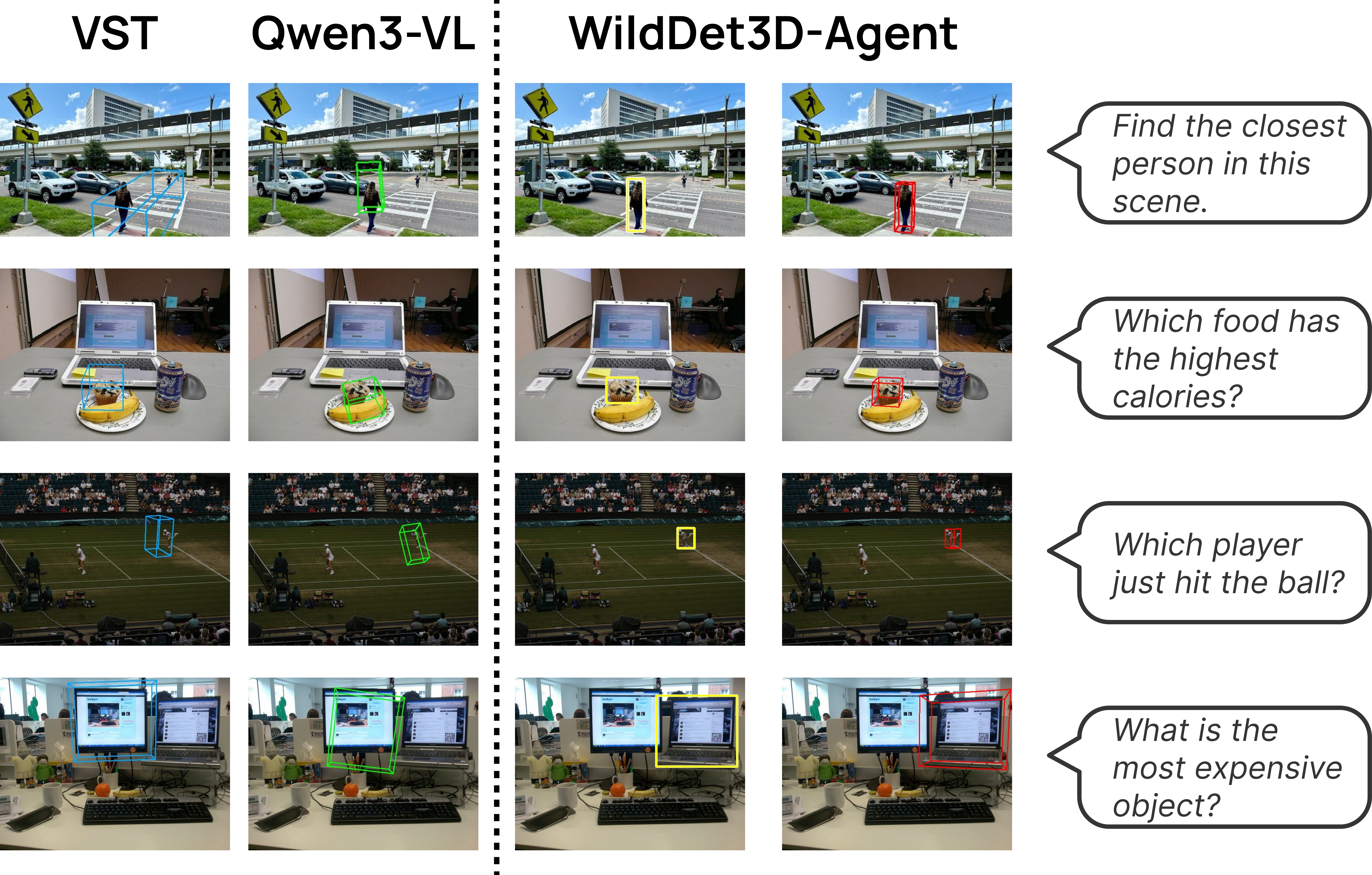}
  \caption{\textbf{\model-agent: referring expression localization.} Results of 3D box outputs by \model compared to VST~\cite{vst} and Qwen3-VL~\cite{qwen3technicalreport}. WildDet3D-Agent more reliably localizes the queried object.}
  \label{fig:spatial_vqa}
\end{figure}

\paragraph{\model-agent: referring expression localization.}
Existing vision-language models can ground objects in 2D, but many real-world spatial questions---``what can I reach from here?'' or ``which object is blocking the door?''---demand 3D understanding that 2D boxes alone cannot provide.
We address this by pairing \model with off-the-shelf grounded VLMs in a two-stage pipeline (Figure~\ref{fig:spatial_vqa}).
Given a free-form query, the VLM performs open-ended reasoning and returns a 2D bounding box around the relevant object; \model then accepts this box as a geometric prompt and lifts it to a full 3D bounding box with metric depth, dimensions, and orientation.
For example, when asked to ``locate the most expensive object in this scene,'' the VLM correctly reasons and recognizes that the computer is probably most expensive and grounds it with a 2D box, after which \model produces the corresponding 3D cuboid. When VLM models such as VST or Qwen3-VL are asked to directly produce the 3D box, they both provided (rather inaccurate) boxes for the monitor instead of the computer. \model-agent can fully leverage language models' visual reasoning capabilities combined with \model's 3D detection ability.
Because \model's box-prompt interface is model-agnostic, any VLM with grounding capability---Molmo~2~\cite{clark2026molmo2openweightsdata}, Qwen3~\cite{qwen3technicalreport}, or future models---can be used as the reasoning front-end in a plug-and-play fashion, turning \model into a universal 3D lifting module that bridges high-level language reasoning with precise spatial localization.

\section{Related work}
\label{sec:related_work}

\paragraph{Monocular 3D object detection.}
Monocular 3D object detection aims to recover 3D bounding boxes from a single RGB image, a fundamentally ill-posed problem due to scale ambiguity, occlusion, and missing geometric cues. Early work in this area largely focused on closed-set settings and domain-specific benchmarks such as autonomous driving~\cite{brazil2019m3d,liu2020smoke,song2024robustness,zhang2023monodetr,wang2021fcos3d,reading2021categorical} and indoor scene understanding~\cite{rukhovich2022imvoxelnet,lazarow2025cubify,7298655,dai2017scannet}. Omni3D~\cite{brazil2023omni3dlargebenchmarkmodel} took an important step toward unification by introducing a cross-dataset benchmark and model spanning multiple indoor and outdoor domains, while follow-up efforts such as UniMODE~\cite{Li_2024_CVPR} further improved unified monocular 3D detection across diverse scenarios. More recent methods have improved geometric reasoning and cross-domain transfer, but most still operate in restricted label spaces or assume a fixed interaction mode.
A recent line of work begins to extend monocular 3D detection toward open-set or open-vocabulary scenarios. Open Vocabulary Monocular 3D Object Detection~\cite{yao2025openvocabularymonocular3d}, 3D-MOOD~\cite{yang20253dmoodlifting2d3d}, and OVM3D-Det~\cite{huang2024training} explore lifting open-vocabulary 2D detections into 3D and show promising generalization beyond fixed category vocabularies. LocateAnything3D~\cite{man2026locateanything3dvisionlanguage3ddetection} further pushes this direction by coupling vision-language models with a chain-of-sight reasoning process for 3D detection. Other methods, such as DetAny3D~\cite{zhang2025detect3dwild}, emphasize promptable 3D box prediction from localized 2D regions. These approaches establish strong baselines, but they typically specialize to a single prompt interface or supervision pipeline: text-query methods are well suited for category-level retrieval, while box-conditioned methods assume oracle or externally provided 2D localization. In contrast, we target a more general setting in which a user may specify an object by text, a point click, or a 2D box, and where additional geometric signals such as sparse depth may be available at test time.

\paragraph{Open-vocabulary and promptable visual perception.}
Our work is also related to the rapid progress in open-vocabulary 2D perception driven by vision-language pretraining. Early grounded and open-vocabulary detectors such as GLIP~\cite{li2022grounded}, OWL-ViT~\cite{minderer2022simple,minderer2023scaling}, and Grounding DINO~\cite{liu2023grounding} showed that large-scale language supervision can support detection beyond fixed taxonomies. In parallel, promptable segmentation systems such as SEEM~\cite{zou2023segment} and SAM 3~\cite{carion2025sam3segmentconcepts} moved visual perception toward a more interactive interface, supporting textual and geometric prompts within a unified framework. 
Recent multimodal LLMs have also pushed visual grounding toward more flexible language-conditioned interaction, including systems for reasoning-based segmentation and pointing such as LISA family~\cite{lai2023lisa,yang2023lisa++,bai2024one} and the Molmo family~\cite{molmov1, clark2026molmo2openweightsdata,clark2025molmopoint}.
We build on this trend, but move from 2D grounding and segmentation to 3D detection: rather than only predicting 2D regions, our model must infer metric center, extent, and orientation in 3D.
More broadly, our work connects to interactive perception systems in which users communicate targets through flexible prompts. Existing 2D systems already support text, clicks, masks, or boxes, but comparable flexibility is rare in monocular 3D detection. Prior 3D systems usually expose either category queries or externally provided 2D boxes, which makes them less suitable for real-world human-in-the-loop applications such as robotics, AR/VR, and grounded visual question answering. Our goal is to bring the prompt flexibility of modern 2D foundation models into 3D, while preserving open-vocabulary recognition and enabling graceful improvement when depth is available.

\paragraph{3D annotation pipelines and open-world 3D data.}
A major bottleneck for generalized 3D detection is data. Compared with 2D detection, large-scale 3D box annotation is substantially more expensive because it requires metric structure, camera parameters, and careful geometric verification. Omni3D~\cite{brazil2023omni3dlargebenchmarkmodel} provides a valuable benchmark across multiple datasets, but it still covers a limited category vocabulary and does not fully reflect the diversity of open-world imagery. More recent efforts therefore explore scalable ways to construct broader supervision, including synthetic data composition for 2D detection and grounding~\cite{huang2026syntheticobjectcompositionsscalable}. In the 3D setting, LabelAny3D~\cite{yao2026labelany3dlabelobject3d} introduces an analysis-by-synthesis pipeline for producing 3D box annotations in the wild and builds COCO3D, a benchmark for open-vocabulary monocular 3D detection. Related efforts such as 3D-MOOD~\cite{yang20253dmoodlifting2d3d} and SAM-3D~\cite{sam3dteam2025sam3d3dfyimages} further demonstrate that lifting 2D cues into 3D, or combining reconstruction with model- and human-in-the-loop annotation, can provide useful supervision at scale. However, automatic annotations remain noisy, especially for object scale, rotation, and extent. Our dataset construction pipeline builds on this insight: instead of relying on a single lifting method, we combine multiple complementary candidate generators and then apply VLM-based scoring, human selection, and geometry-aware filtering to obtain a large-scale in-the-wild dataset for open-vocabulary 3D detection.

\section{Limitations}
\label{sec:limitations}

While \model achieves strong results across diverse settings, several limitations remain.

\paragraph{Camera intrinsics accuracy.}
Our geometry backend can predict camera intrinsics when they are not provided, enabling fully uncalibrated inference. However, predicted intrinsics are less accurate than ground-truth calibration, leading to degraded 3D localization, particularly for absolute depth and physical dimensions. Closing this gap remains an open challenge for in-the-wild deployment where camera metadata is unavailable.

\paragraph{Single-image depth ambiguity.}
Monocular 3D detection is inherently ill-posed: a single image cannot fully resolve metric depth without additional cues. Our geometry backend mitigates this through learned depth priors, but performance on distant or heavily occluded objects remains limited. The substantial gains from sparse depth input (Table~\ref{tab:omni3d}) highlight this fundamental bottleneck.

\paragraph{Rotation estimation.}
Despite unambiguous rotation normalization, rotation prediction remains the weakest component of our 3D box estimation. Objects with near-symmetric geometry (\eg, round tables, square boxes) or limited visible surface area pose particular challenges, as the visual signal for orientation is inherently ambiguous.

\paragraph{Computational cost.}
The dual-backbone design (vision encoder + geometry backbone running in parallel) increases memory and compute requirements compared to 2D-only detectors. While acceptable for server-side deployment, the full model is too large for real-time on-device inference without distillation or quantization.

\paragraph{Long-tail categories.}
Performance on rare categories in open-world evaluation lags behind frequent ones. The long-tailed distribution of \data partially addresses this, but categories with very few training examples still exhibit high variance in 3D prediction quality.

\paragraph{Intended use and deployment boundaries.}
The applications demonstrated in Section~\ref{sec:application} (iPhone, AR, robotics, VLM integration) are intended as research prototypes illustrating the versatility of open-vocabulary 3D detection, not as production-ready systems. Predictions may contain incorrect depth, dimensions, or missed detections, and the model provides no guaranteed error bounds. \model is \emph{not intended for safety-critical applications} such as autonomous navigation, surgical planning, or structural assessment.

\section{Conclusion}
\label{sec:conclusion}

We presented \model, an open-vocabulary monocular 3D object detector that unifies text, point, and box prompts within a single geometry-aware architecture, and \data, a large-scale in-the-wild dataset spanning 1M images and 13.5K categories with human-verified 3D annotations.

On the model side, \model introduces dual vision encoders with a depth fusion module that gracefully incorporates optional depth input, an integrator that accommodates diverse prompt modalities, and a 3D detection head that aggregates depth, spatial, and semantic features.
On the data side, \data expands category coverage by 138$\times$ over Omni3D through a multi-model candidate generation pipeline followed by two-stage human and VLM selection, providing broad open-world supervision previously unavailable for 3D detection.

Experiments demonstrate that \model achieves state-of-the-art results on Omni3D (34.2 AP$_\text{3D}$ text, 36.4 AP$_\text{3D}$ oracle) with 6--10$\times$ fewer training epochs than prior methods, generalizes zero-shot to Argoverse~2 and ScanNet (40.3 and 48.9 ODS), and shows strong open-world transfer across 700+ in-the-wild categories.
We further demonstrate practical deployment on iPhone, Meta Quest 3, robotic manipulation, and VLM-based spatial reasoning, showing that \model serves as a general-purpose 3D perception module across diverse platforms and applications.

\section*{Acknowledgements}

This work would not be possible without the support of our colleagues at Ai2. We thank David Albright, Kristin Cha, Stephen Kelman, Yiqin Dai, Byron Bischoff, Cailin Brashear, Caleb Ouellette, David Everhart, Emily Mullen, Jon Borchardt, Crystal Nam, Patricia Balik, Tina Weiss, Kyle Wiggers, Will Smith, Peter Clark, and Noah Smith for their important work for the \model public release.

\clearpage
\bibliographystyle{abbrvnat}
\bibliography{main}

\clearpage
\appendix

\section*{Appendix}

The appendix includes the following sections:
\begin{itemize}
\itemsep0em
    \item \S\ref{supp:model} - Model and loss details
    \item \S\ref{supp:training} - Training details
    \item \S\ref{appendix:eval} - Evaluation details
    \item \S\ref{appendix:data} - Dataset details
    \item \S\ref{appendix:qualitative_dataset} - Dataset examples
    \item \S\ref{appendix:qualitative} - Qualitative results
\end{itemize}

\section{Model and loss details}
\label{supp:model}

All geometry backend losses are scaled by a global factor $\lambda_\text{geom}\!=\!5.0$.
Each component is clipped to a maximum of~10 before scaling to prevent gradient explosion from outlier pixels.

\subsection{Auxiliary geometry loss \texorpdfstring{$\mathcal{L}_\text{geom}$}{L\_geom}}

The geometry backend produces depth estimates, 3D point maps, a confidence mask, and camera intrinsics.
The auxiliary geometry loss comprises eight terms:

\paragraph{Metric depth L1.}
A standard L1 loss between predicted and ground-truth depth at valid pixels (where $d^*\!>\!0$ and depth ratio $\hat{d}/d^*\!\in\![1/3,\,3]$):
\begin{equation}
  \mathcal{L}_\text{depth-L1} = \frac{1}{|\mathcal{V}|}\sum_{p\in\mathcal{V}} |\hat{d}_p - d_p^*|,
\end{equation}
where $\mathcal{V}$ is the set of valid pixels.
Weight: $w\!=\!1.0$.

\paragraph{Scale-invariant logarithmic depth (SILog).}
Following Eigen~\etal~\cite{eigen2014depthmappredictionsingle}:
\begin{equation}
  \mathcal{L}_\text{SILog} = \sqrt{\text{Var}(g) + 0.15\cdot\text{Mean}(g)^2}, \quad g_p = \log\hat{d}_p - \log d_p^*.
\end{equation}
Weight: $w\!=\!0.5$.

\paragraph{Affine-invariant point-map losses.}
We back-project predicted and ground-truth depth maps to 3D point clouds and compute three MoGe2-based~\cite{wang2025moge2accuratemonoculargeometry} losses:
\begin{itemize}[nosep]
  \item \emph{Global alignment} ($w\!=\!10.0$): aligns the predicted point cloud to the GT via optimal affine transform at resolution~$48^2$;
  \item \emph{Local alignment} at two scales ($w\!=\!10.0$ each): level-4 ($24^2$ patches, 16 samples) and level-16 ($12^2$ patches, 256 samples), capturing fine-grained local geometry;
  \item \emph{Edge loss} ($w\!=\!10.0$): penalizes depth discontinuity mismatches at object boundaries.
\end{itemize}

\paragraph{Confidence mask BCE.}
A per-pixel binary cross-entropy loss that supervises the depth validity confidence prediction against a three-state ground-truth mask (finite depth, infinite/invalid depth, and unknown).
For sparse depth inputs (coverage~$<70\%$), only annotated pixels contribute to the loss.
Weight: $w\!=\!0.1$.

\paragraph{Camera ray MSE.}
An L2 loss between predicted and ground-truth camera ray directions, derived from the respective intrinsics matrices:
\begin{equation}
  \mathcal{L}_\text{ray} = \text{MSE}(\mathbf{r}(\hat{\mathbf{K}}),\;\mathbf{r}(\mathbf{K}^*)),
\end{equation}
where $\mathbf{r}(\mathbf{K})$ denotes the ray direction field generated from intrinsics~$\mathbf{K}$.
Weight: $w\!=\!1.0$.

\subsection{Auxiliary 2D detection loss \texorpdfstring{$\mathcal{L}_\text{2D}$}{L\_2D}}

The 2D detection losses follow the SAM~3~\cite{carion2025sam3segmentconcepts} design with minor modifications.

\paragraph{IoU-aware classification (BCE).}
For each matched prediction--target pair, the classification target is an IoU-aware soft label:
\begin{equation}
  t = \sigma(z)^{\alpha}\cdot\text{IoU}_\text{2D}^{1-\alpha}, \quad \alpha = 0.25,
\end{equation}
where $\sigma(z)$ is the predicted probability and $\text{IoU}_\text{2D}$ is the 2D box IoU with the matched GT.
Positive predictions are weighted by $w_+\!=\!5$; unmatched predictions receive a focal-weighted negative loss with $\gamma\!=\!2$.
Loss weight: $w\!=\!20$.

\paragraph{Box regression.}
Combines an L1 loss on normalized center-size coordinates ($w\!=\!5$) and a generalized IoU loss~\cite{rezatofighi2019generalizedintersectionunionmetric} on pixel-space boxes ($w\!=\!2$):
\begin{equation}
  \mathcal{L}_\text{box} = 5\cdot\text{L1}(\hat{b}_\text{cxcywh},\,b^*_\text{cxcywh}) + 2\cdot(1 - \text{GIoU}(\hat{b},\,b^*)).
\end{equation}

\paragraph{Per-category presence.}
A sigmoid BCE loss ($w\!=\!20$) that predicts whether each queried category has any instance in the image.
Uses $\alpha\!=\!0.5$ and $\gamma\!=\!0$ (plain BCE without focal weighting).

\paragraph{One-to-many (O2M) matching.}
Each ground-truth box is matched to its top-$k$ ($k\!=\!4$) scoring predictions using a binary matcher with IoU threshold~0.4.
The same classification, box, and 3D losses are computed for all matched pairs, scaled by $w_\text{o2m}\!=\!2.0$ and clipped at~150 to prevent gradient explosion.

\section{Training details}
\label{supp:training}

\subsection{Three-stage training pipeline.}

Table~\ref{tab:training_stages} summarizes the three training stages.
All stages use AdamW with base learning rate $10^{-4}$, weight decay
$10^{-4}$, 4 nodes (32~GPUs), and per-GPU batch size 4 (total 128).
The learning rate follows a multi-step decay: for stages with $N$
epochs, it decays by $0.1\times$ at epochs $\lfloor N \cdot s_1 \rfloor$
and $\lfloor N \cdot s_2 \rfloor$.

\begin{table}[h]
\centering
\caption{\textbf{Training stage summary.}}
\label{tab:training_stages}
\vspace{-0.5em}
\tablestyle{4pt}{1.15}
\small
\begin{tabular}{l | c c c c}
\toprule
Stage & Data & Epochs & LR decay ($s_1$/$s_2$) & Init \\
\midrule
1 & Omni3D & 12 & $2/3$ / $5/6$ & Scratch \\
2 & Omni3D + Others + \data (H+S) & 12 & $2/3$ / $5/6$ & Stage~1 \\
3 & Omni3D + \data (H) & 3 & $1/3$ / $2/3$ & Stage~2 \\
\bottomrule
\end{tabular}
\end{table}

\paragraph{Stage 2 data mixing ratios.}
Stage~2 combines seven datasets with the following sampling proportions:
Omni3D 40\%, CA-1M 10\%, Waymo 5\%, 3EED-det 2.5\%, 3EED-ref 2.5\%,
FoundationPose 20\%, and \data (human + synthetic) 20\%.

\paragraph{Stage 3 mask-guided training.}
Stage~3 uses Omni3D (90\%) and \data human annotations (10\%).
To leverage 2D segmentation masks from SAM~2~\cite{ravi2024sam2}, we apply
mask-guided point/box training: for images with available masks,
box prompts and point prompts sampled inside the mask region are
used as geometric inputs, encouraging the model to learn tighter
3D localization from precise 2D evidence.

\paragraph{Freeze configuration.}
The SAM3 ViT backbone has its first 28 transformer blocks frozen
across all stages.
The LingBot-Depth geometry backend uses a ViT-L encoder (24 blocks);
the first 21 blocks are frozen and the last 3 remain trainable to
allow the depth encoder to adapt to new data distributions.
The 3D detection head is trained from scratch in all stages.

\section{Evaluation details}
\label{appendix:eval}

We describe the evaluation metrics and protocols used across all benchmarks.

\paragraph{3D IoU matching (bbox mode).}
For Omni3D evaluation, we use 3D bounding box IoU as the matching criterion.
Predictions are matched to ground truths per category using 3D IoU, computed via oriented bounding box overlap from the 8 corner points in camera coordinates~\cite{brazil2023omni3dlargebenchmarkmodel}.
AP is averaged over 10 IoU thresholds $\tau \in \{0.05, 0.10, \ldots, 0.50\}$, with per-threshold AP values denoted AP15, AP25, AP50 for $\tau = 0.15, 0.25, 0.50$.
We also report results stratified by object depth: near ($<$10\,m), medium (10--35\,m), and far ($>$35\,m).
Maximum detections per image is capped at 100.

\paragraph{Center-distance matching (dist mode).}
For WildDet3D-Bench and Stereo4D, we use center-distance matching.
A prediction is matched to a ground truth if the Euclidean distance between their 3D centers is below a threshold proportional to the object's spatial extent:
\begin{equation}
  \|\hat{\mathbf{c}} - \mathbf{c}^*\| < \tau \cdot r, \quad r = \frac{\|\mathbf{d}^*\|_2}{2},
\end{equation}
where $\mathbf{d}^* = (w, h, l)$ are the GT dimensions and $r$ is the object half-diagonal (radius).
AP is averaged over 11 distance thresholds $\tau \in \{0.50, 0.55, \ldots, 1.00\}$.
Depth stratification follows the same near/medium/far splits.

\paragraph{Open Detection Score (ODS).}
For zero-shot evaluation on Argoverse~2 and ScanNet, we report the Open Detection Score~\cite{yang20253dmoodlifting2d3d}, a composite metric:
\begin{equation}
  \text{ODS} = \frac{3 \cdot \text{AP} + (1 - \text{mATE}) + (1 - \text{mAOE}) + (1 - \text{mASE})}{6},
\end{equation}
where mATE is the mean translation error (center distance normalized by the matching distance threshold), mAOE is the mean absolute orientation error (normalized by $\pi$), and mASE is the mean scale error ($1 - \text{IoU}_\text{scale}$, where $\text{IoU}_\text{scale}$ is the volumetric IoU computed from axis-aligned dimension overlap).
AP contributes 50\% of the score (weight 3/6), encouraging both detection quality and geometric accuracy.

\paragraph{Frequency-split AP.}
On WildDet3D-Bench, we partition the 700+ evaluation categories into three groups based on the number of images containing each category: rare ($<$5), common (5--20), and frequent ($>$20), and report per-group AP (AP$_r$, AP$_c$, AP$_f$).
This follows the LVIS~\cite{gupta2019lvis} evaluation protocol for long-tail category distributions.

\paragraph{Federated evaluation.}
Since WildDet3D-Bench annotations are not exhaustive---not every object in each image has a valid 3D bounding box---we follow the federated evaluation protocol of LVIS~\cite{gupta2019lvis}: a prediction that overlaps with a 2D-annotated object lacking a valid 3D box is treated as neutral rather than a false positive.

\paragraph{Post-processing.}
At test time, we apply per-category NMS with a 2D IoU threshold of 0.6.
Predictions with a 2D objectness score below 0.05 are discarded before evaluation.

\section{Dataset details}
\label{appendix:data}

\paragraph{Annotator screening.}
Before participating in the main annotation study, workers on
Prolific complete a 10-task screening batch ($\sim$5 minutes,
\$1.50 reward).  The screening tasks are drawn from a curated gold
set and assess two independent skills:
(1)~\textbf{Unacceptable detection} (Filter~1): annotators must
correctly flag $\geq$\,2/3 known-bad annotations as
\texttt{unacceptable} without mislabeling any as \texttt{good\_fit};
(2)~\textbf{Candidate selection accuracy} (Filter~2): annotators
must select the correct best candidate on $\geq$\,5/7 candidate
tasks.  Both filters must be passed to qualify.  A total of 500
screening batches were issued; qualified workers are then invited to
the main annotation batches.

\paragraph{3D box translation optimization.}
Each candidate box is aligned to the scene depth via a two-stage
optimization.  First, a coarse grid search evaluates $5^3 = 125$
candidate translations within a window proportional to the box
dimensions (scale factor 1.0), using a combined loss:
\begin{equation}
  \mathcal{L} = \mathcal{L}_\text{3D} + \lambda_\text{2D}\,\mathcal{L}_\text{2D},
  \quad \lambda_\text{2D} = 0.5
\end{equation}
where $\mathcal{L}_\text{3D} = \lambda_\text{in}\,\mathcal{L}_\text{in} +
\lambda_\text{tight}\,\mathcal{L}_\text{tight}$
($\lambda_\text{in}{=}1.0$, $\lambda_\text{tight}{=}0.5$)
combines an \emph{inclusion loss} (anchor points should lie inside
the box, buffer 0.02\,m) and a \emph{tightness loss} (box faces
should be close to point cloud, buffer 0.1\,m);
$\mathcal{L}_\text{2D} = 1 - \text{GIoU}(\hat{b}_{2D}, b_{2D})$
penalizes 2D projection mismatch.
The best grid point initializes an L-BFGS-B local optimizer
(max 100 iterations, $f$-tolerance $10^{-6}$).
Box dimensions and rotation are kept fixed throughout; only the
3D center is optimized.
An adaptive IoU switch (threshold 0.4) selects between this
optimized result and a simpler height-based scaling method
($0.7\times$height $+ 0.3\times$width scale).
Anchor points (256 per box) are sampled from an eroded mask
point cloud with Mahalanobis-distance weighting
($\alpha{=}0.5$) to downweight outliers.

\paragraph{Val/test sampling.}
The three-phase balanced sampling targets the following
per-split distributions.
\textbf{Depth} (object-level):
near ($<$10\,m) 50\%,
mid (10--35\,m) 25\%,
far (35--100\,m) 20\%,
super-far ($>$100\,m) 5\%.
\textbf{Source}: COCO 20\%, LVIS 40\%, Objects365 40\%.
Categories with fewer than 3 sampled images are marked as
\texttt{rare\_category} and excluded from evaluation.

\paragraph{Small object upgrade.}
Objects initially filtered as small (2D area $<$\,0.5\% of image)
are re-evaluated in a separate pipeline.
A candidate qualifies if: VLM score $\geq$\,10, category
sub-score = 1, and model $\in$ \{LabelAny3D, SAM-3D,
RANSAC-PCA\}.  Candidates with VLM score = 10 additionally
require 3D-to-2D projected IoU $\geq$\,0.5 (score 11 is exempt).
Per-category selection is capped at 1{,}500 annotations,
grouping images together to avoid splitting same-image annotations.

\paragraph{GPT-4.1-mini category size estimation.}
For each object category, GPT-4.1-mini (temperature 0) is prompted
to estimate the physical 3D bounding box dimensions in metres,
returning six range fields (shortest/middle/longest axis min/max)
plus Boolean flags \texttt{is\_flat} and \texttt{is\_elongated}.
The system prompt instructs the model to be generous with ranges
(catching clearly wrong sizes, not borderline cases), and to
return a JSON object only.  Flat categories (flags, plates,
posters) skip the shortest-axis check; elongated categories
(poles, pens, bats) skip the shortest and longest axes.

\begin{figure*}[t]
  \centering
  \includegraphics[width=0.49\textwidth]{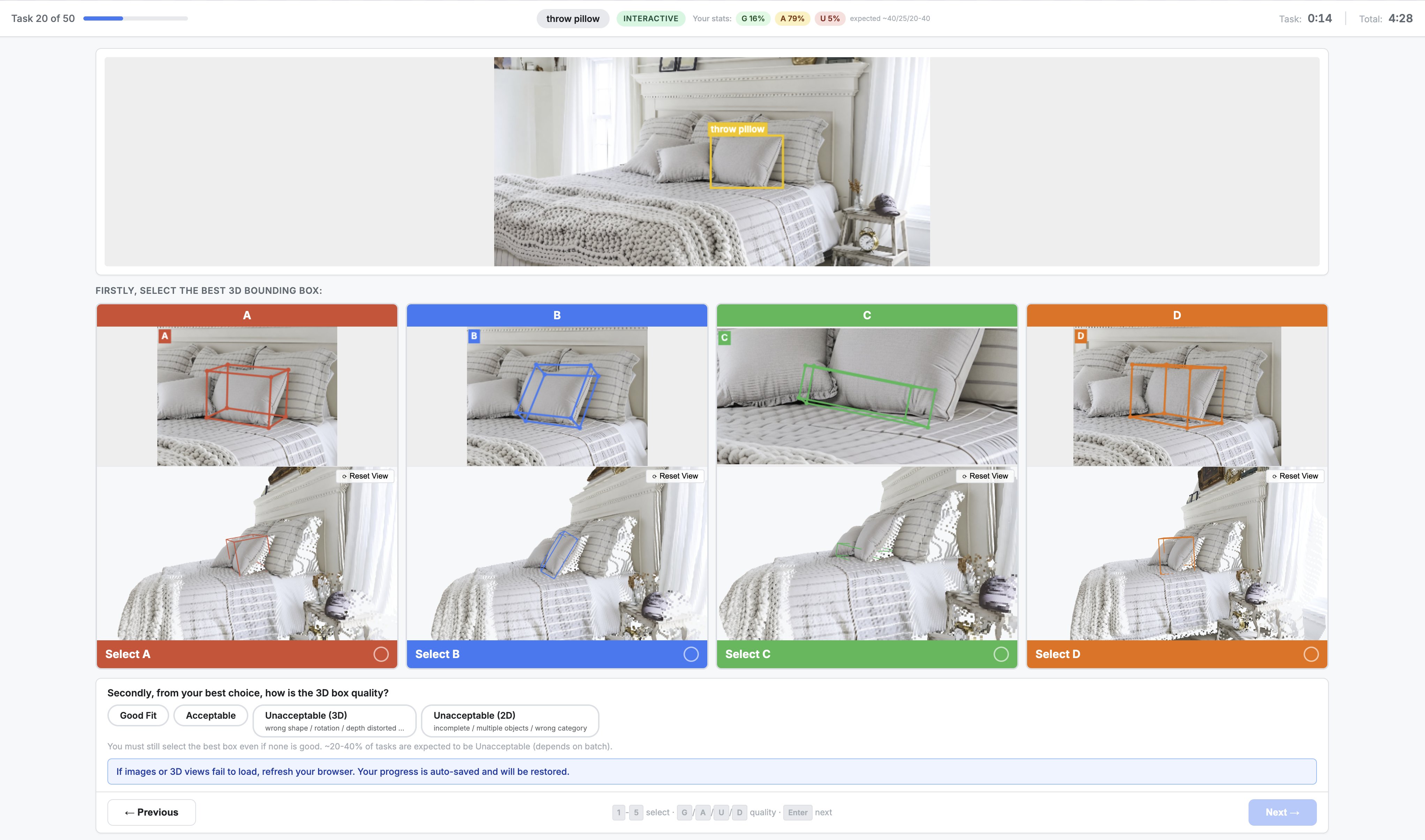}
  \hfill
  \includegraphics[width=0.49\textwidth]{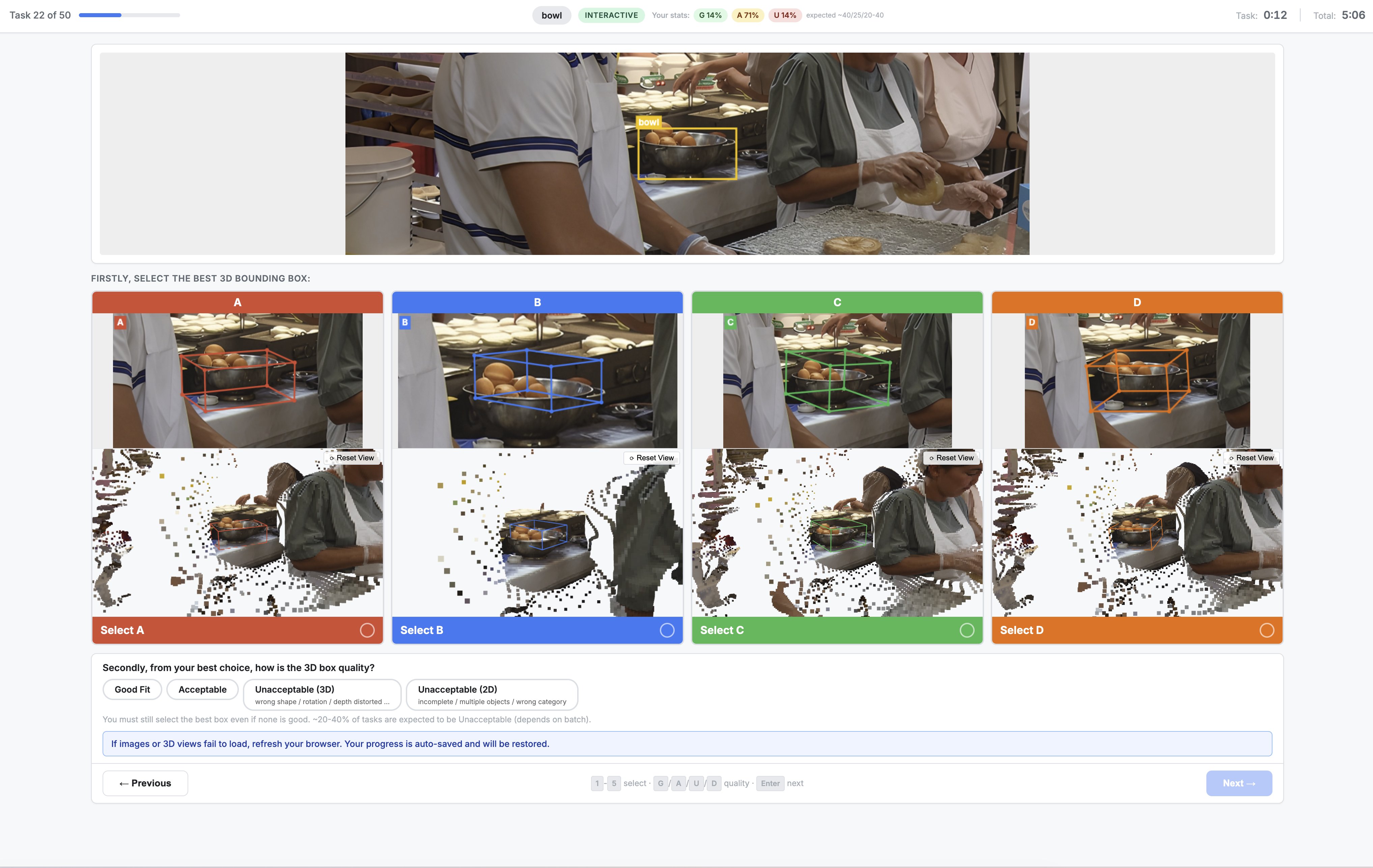}
  \caption{\textbf{Annotation interface.}
  Two example tasks from the \data annotation interface.
  The top panel shows the reference image with the target 2D box
  highlighted in yellow and all candidate 3D boxes projected in
  color.  Each candidate column (A--D, color-coded) shows a cropped
  2D projection and an orthographic point cloud view.
  Annotators select the best candidate and rate its quality as
  \texttt{good\_fit}, \texttt{acceptable}, \texttt{unacceptable (3D)},
  or \texttt{unacceptable (2D)}.}
  \label{fig:annotation_interface}
\end{figure*}

\paragraph{Annotator demographics.}
A total of 1,786 unique annotators participated through Prolific.
The pool is gender-balanced (50.7\% male, 49.0\% female), with ages
ranging from 18 to 90 (mean 42.1).  Annotators are predominantly
from English-speaking countries: United States (55.9\%), United
Kingdom (31.8\%), and Canada (7.0\%), consistent with the English
language requirement of the task.  Ethnicity is distributed as
White (78.0\%), Black (9.4\%), Asian (5.7\%), and Mixed (4.9\%).

\paragraph{Ethics.}
Annotation tasks were conducted on the Prolific platform. Participants voluntarily accepted tasks and were compensated at \$3.50 per batch of 55 annotations (average completion time 12--15 minutes, corresponding to an effective hourly rate of \$14--17.50), meeting or exceeding the minimum wage standards of all participating countries. Demographic statistics are aggregate data provided by the platform. All studies were reviewed and approved by Prolific's platform guidelines to ensure fair treatment and ethical standards for participants.
The annotator pool skews toward English-speaking Western countries (86\% US/UK/CA, 78\% White), which may influence judgments on what constitutes a plausible 3D box for culture-specific object categories or unfamiliar scene layouts. Similarly, the LLM-based size filters and VLM scoring heuristics used in the pipeline inherit biases from their training data, potentially affecting which annotations are retained or rejected for underrepresented object types or scenes. We report these demographics and pipeline choices transparently so that downstream users can account for potential biases.


\section{More \data examples}
\label{appendix:qualitative_dataset}

\begin{figure*}[h]
  \centering
  \includegraphics[width=\textwidth]{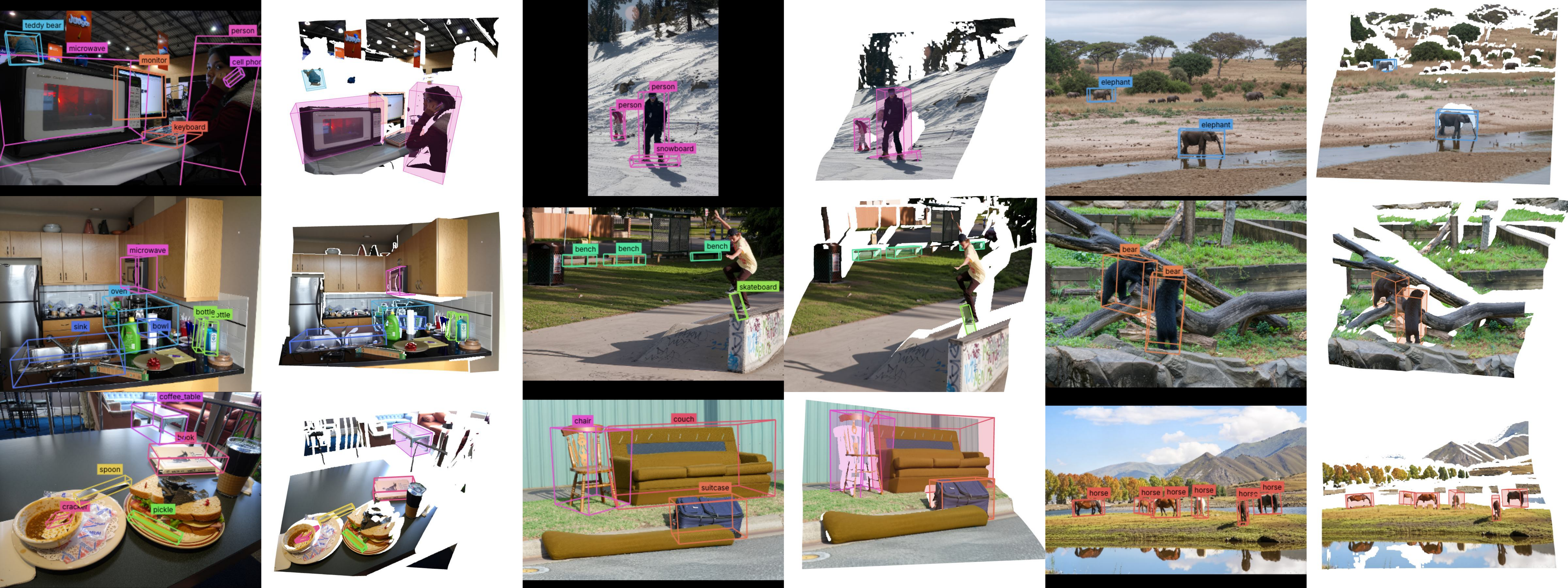}
  \caption{\textbf{Qualitative examples from \data.}
  Each pair shows 3D bounding box annotations overlaid on the input image with category labels (left) and the corresponding 3D bounding boxes rendered in the reconstructed point cloud (right).}
  \label{fig:qualitative_data_appendix}
\end{figure*}

See Figure~\ref{fig:qualitative_data_appendix} for additional examples.

\newpage
\section{Additional qualitative results}
\label{appendix:qualitative}

\begin{figure*}[h]
  \centering
  \includegraphics[width=\textwidth]{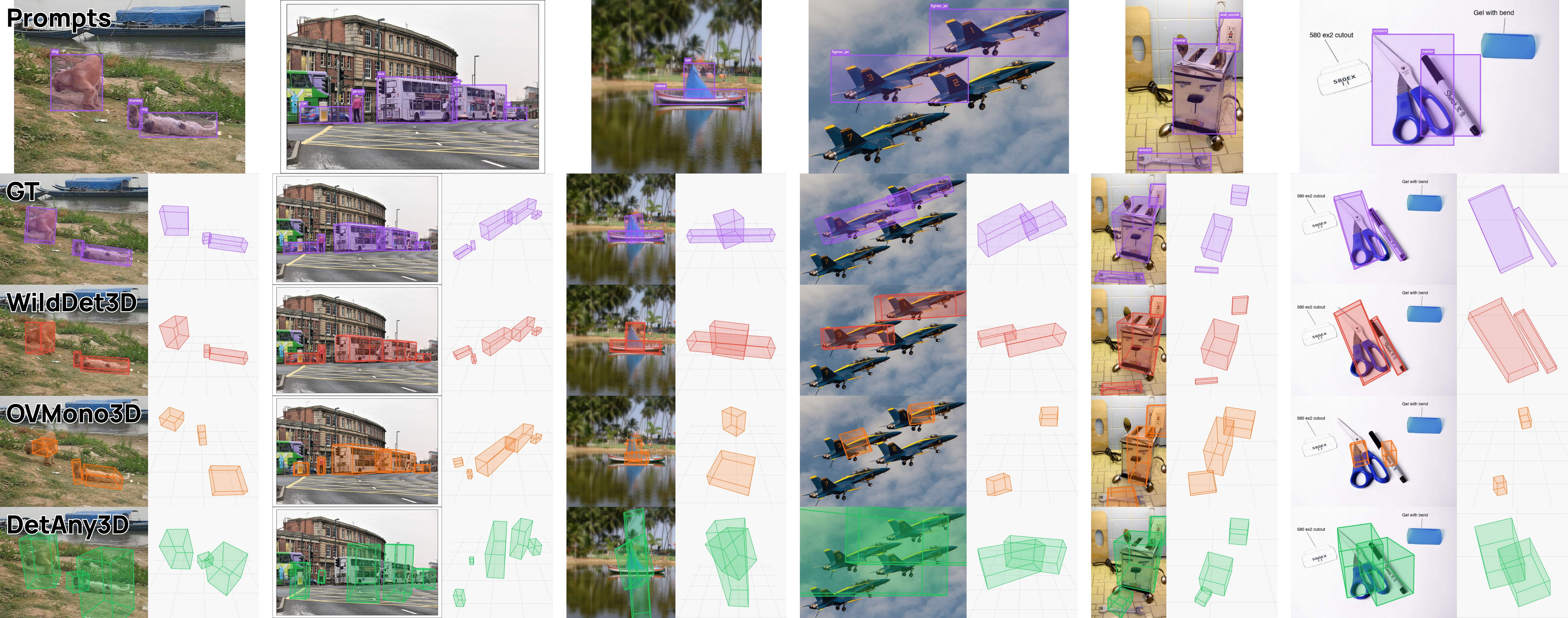}
  \caption{\textbf{Box-prompted comparison.}
  Each block shows the same scene detected by three models, all prompted using 2D bounding boxes. From top to bottom: 2D box prompt visualizations (only box prompts are used, the text labels are for reference), ground truth 3D boxes, \model predictions, OVMono3D predictions, and DetAny3D predictions, with 2D overlays and corresponding 3D bounding boxes.}
  \label{fig:model_comparison_box_appendix}
\end{figure*}

\begin{figure*}[h]
  \centering
  \includegraphics[width=\textwidth]{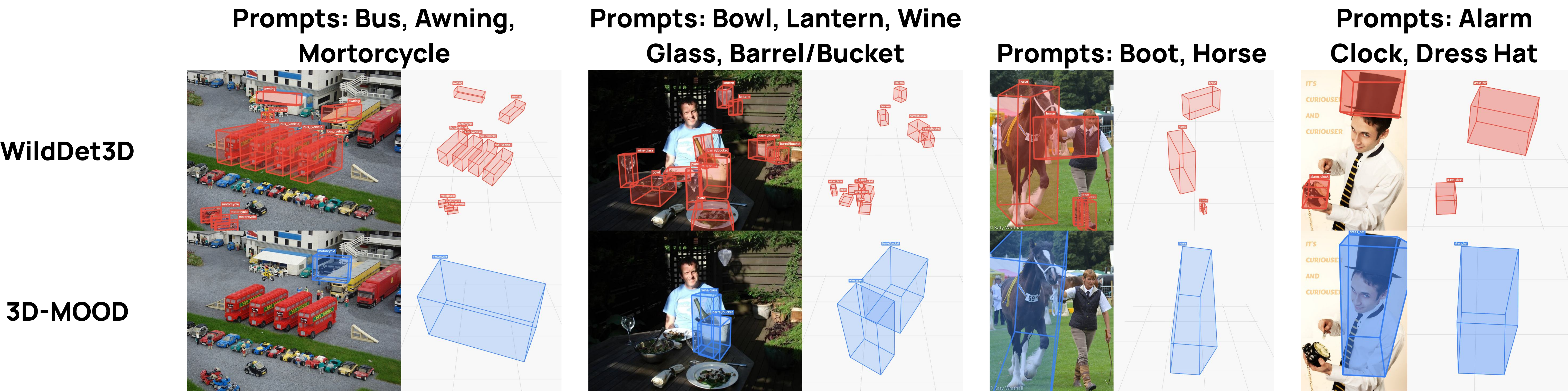}
  \caption{\textbf{Text-prompted comparison.}
  Each block shows the same scene detected by WildDet3D (top) and 3D-MOOD (bottom), prompted with text categories only.}
  \label{fig:model_comparison_text_appendix}
\end{figure*}

See Figure~\ref{fig:model_comparison_box_appendix} and Figure~\ref{fig:model_comparison_text_appendix} for additional examples.

\end{document}